\documentclass[preprint,12pt]{elsarticle}
\usepackage{newtxtext}

\usepackage{lineno}

\usepackage[bottom]{footmisc} 

\usepackage{geometry}
\usepackage{amsmath}
\usepackage{algorithm}
\usepackage{algpseudocode}
\usepackage{url}
\usepackage{upgreek}
\usepackage{pdflscape}
\usepackage{caption}
\usepackage{setspace}
\usepackage{pifont}
\usepackage{amsmath}
\usepackage{bbding}
\usepackage{amsfonts} 
\usepackage{amssymb}
\usepackage{mathrsfs}

\usepackage{color}
\usepackage{dsfont}
\usepackage[misc]{ifsym}

\usepackage{graphicx}  
\usepackage{array}
\usepackage{verbatim}
\usepackage{titletoc}
\usepackage{color,xcolor}

\usepackage{units}
\usepackage{multirow}
\usepackage{tabularx}
\usepackage{booktabs}
\usepackage{bbm}
\usepackage{amsmath}     
\usepackage{bm} 
\usepackage{algorithm}
\usepackage{algpseudocode}
\usepackage{fvextra}

\definecolor{mediumgray}{gray}{0.88}
\definecolor{lightgray}{gray}{0.95}

\usepackage{colortbl}

\usepackage{amsmath}

\usepackage{hyperref}
\hypersetup{
    colorlinks=true,
    linkcolor=blue,
    citecolor=blue,
    urlcolor=blue
}

\usepackage{xcolor}
\definecolor{DeBlue}{RGB}{68,114,196}
\definecolor{DeGreen}{RGB}{84,130,53}
\definecolor{DeBlack}{RGB}{0,0,0}

\usepackage{natbib}
\usepackage{fancyhdr}

\begin{document}

\title{PARA-PV: Physics-Aware Retrieval-Augmented PV Prediction Based on Frozen Foundation Model and Distribution Shift Correction} 

\begin{frontmatter}

\author[inst1]{Hang Fan}
\affiliation[inst1]{organization={School of Economics and Management},
            addressline={North China Electric Power University}, 
            city={102206, Beijing},
            country={China}}

\author[inst2]{Weican Liu\corref{cor1}}
\affiliation[inst2]{organization={School of Electrical and Electronic Engineering},
             addressline={Nanyang Technological University}, 
             city={639798, 50 Nanyang Avenue},
             country={Singapore}}   
\cortext[cor1]{W. Liu is the Corresponding author (Email: weican001@e.ntu.edu.sg).}

\author[inst1]{Ying Lu} 

\author[inst1]{Dunnan Liu}

\author[inst3]{Long Cheng}
\affiliation[inst3]{organization={School of Control and Computer Engineering},
            addressline={North China Electric Power University}, 
            city={102206, Beijing},
            country={China}}

\author[inst4]{Wei Wei\corref{cor2}}
\cortext[cor2]{W. Wei is the Co-corresponding author (Email: wei-wei04@mails.tsinghua.edu.cn).}
\affiliation[inst4]{organization={Department of Electrical Engineering},
             addressline={Tsinghua University},
             city={100084, Beijing},
            country={China}}

\normalsize

\begin{abstract}
Accurate photovoltaic (PV) power forecasting is essential for reliable grid dispatch and renewable energy integration, yet it remains challenging because PV generation is jointly shaped by weather variability, day-night transitions, regime-dependent dynamics, and strict physical constraints. Existing data-driven methods typically improve temporal representation learning by increasing model complexity, but they insufficiently exploit PV-specific physical knowledge and may produce biased forecasts under peak, ramping, or low-power conditions. To address these limitations, we propose PARA-PV, a Physics-Aware Retrieval-Augmented framework that embeds physical knowledge throughout the forecasting process. The framework first encodes multivariate PV observations into patch-level representations and, through a physics-aware retrieval-augmented learner, retrieves historical patches and analog trajectories that are consistent with the current window in temporal shape, power level, PV operating state, and intra-day period; this yields a physically grounded base forecast. To supplement local memory with broader temporal knowledge, the base forecast is then calibrated against a frozen Chronos time-series foundation-model prior through a lightweight residual adapter, so that general temporal regularities are adapted to PV-specific dynamics without overriding the physically grounded prediction. Because residual conditional distribution shifts persist when weather and diurnal regimes change, a physics-aware distribution shift correction module subsequently adjusts the preliminary forecast using power, weather, timestamp, and day/night conditions, applying gated mean-shift and scale corrections selectively. Finally, a physics-constrained loss function partitions the samples into peak, ramping, night-time, and regular regimes and adaptively reweights their error contributions, preventing the dominant regular regime from suppressing learning of operationally critical states. Experiments on two geographically distinct PV farms demonstrate that PARA-PV consistently outperforms seven representative baselines across multiple forecasting horizons, achieving lower point-forecast errors and sharper prediction intervals with fewer than 1M trainable parameters. Our code is available at \url{https://github.com/weican1103/PARA-PV}.
\end{abstract}


\end{frontmatter}
\setcounter{page}{1}

\section{Introduction}

\subsection{Background and motivation}

With the continuous growth of global energy demand, the development of reliable and low-carbon energy systems has become an urgent priority. Among renewable energy sources, solar energy offers clear advantages in terms of cleanliness, scalability, and wide geographical availability. Driven by advances in PV technology and policy support, PV generation has expanded rapidly and is becoming an increasingly important component of modern power systems \cite{FAN2026104455}. Accurate PV power forecasting is essential for grid integration of solar power and energy management  \cite{HUANG2026127816, XIA2026127331, TIAN2025125525, WANG2025137654, HUANG2025124753}. However, PV forecasting remains challenging because power output is jointly affected by multiple meteorological and temporal variables, such as irradiance, cloud movement, and diurnal cycles, while also being constrained by physical operating regimes including night-time zero generation, capacity limits, peak production, and rapid ramping dynamics, which make it difficult to learn hidden information \cite{YANG2025124744}.

Although PV forecasting models have evolved from statistical methods to deep temporal architectures, their predictive reliability remains limited under complex meteorological and operating conditions. First, many existing models mainly pursue more expressive data-driven architectures, but they insufficiently incorporate the physical constraints of PV generation, such as zero output at night, capacity limits, peak-generation regimes, and rapid ramping behavior. These models may fit historical observations well but still produce physically inconsistent or regime-biased forecasts when weather conditions and operating states change. Second, most studies have not fully exploited recent advances in time-series foundation models, whose general temporal modeling capability can provide useful priors for prediction when properly calibrated. Third, existing approaches often focus on improving the forecasting model itself while paying less attention to post-prediction distribution shift correction. However, meteorological variations and temporal conditions can shift the distribution of PV output. Therefore, simply using weather or time variables as model inputs may improve accuracy, but it does not fundamentally correct the distributional bias of prediction.

\subsection{Related work}

Recently, high-accuracy PV power forecasting models have developed rapidly. According to previous studies, forecasting models can be broadly classified into physical models, statistical models, and artificial intelligence models. Physical models estimate power generation by describing the energy conversion process of PV systems based on solar irradiance, module characteristics, and system configuration. For example, Zhi \textit{et al.} proposed a physical model that combines meteorological parameter prediction, irradiance calculation, an equivalent circuit model and an improved maximum power tracking algorithm for rooftop PV power prediction, achieving a mean absolute error of 15.9\% under various weather conditions \cite{ZHI2023106997}. While these models are generally interpretable and physically consistent, their performance often depends on the accuracy of meteorological inputs and detailed system parameters \cite{WOLFF2016197}. Statistical models, including ARIMA (Autoregressive Integrated Moving Average) \cite{LI201478} and regression-based models \cite{en15114171}, mainly establish mathematical relationships between historical PV power observations and relevant explanatory variables, offering efficient prediction. For example, Doubleday \textit{et al.} proposed a Bayesian model averaging post-processing method for probabilistic utility-scale PV forecasting, which calibrates biased numerical weather prediction ensembles through weighted probability density functions \cite{9091068}. Although statistical models can capture trend variations, they often face challenges in modeling complex nonlinear relationships \cite{AHMED2020109792}.

Driven by recent advances in artificial intelligence, machine learning methods, such as Support Vector Regression (SVR) \cite{das2017svr}, Random Forest (RF) \cite{xie2024rsmd}, and Gradient Boosting Decision Tree (GBDT) \cite{wang2023novel}, have been increasingly applied to time-series forecasting tasks. For example, Eseye \textit{et al.} proposed a Hybrid Wavelet-PSO-SVM model, which integrates wavelet transform and support vector machine to model nonlinear relationships between numerical weather prediction variables and PV power \cite{eseye2018short}. Hategan \textit{et al.} proposed a weighted ensemble model that combines machine learning and sky-image-based methods, improving intra-hour PV power forecasting accuracy \cite{hategan2025short}. Although machine learning models have been widely used for PV power forecasting, their performance is often limited by strong data dependence, weak physical interpretability, and reduced robustness under rapidly changing weather conditions \cite{liu2025pv}. Deep learning models have been increasingly used for PV power forecasting due to their stronger ability to capture nonlinear temporal patterns, such as recurrent neural network-based models \cite{ahn2021deep}, graph convolutional network-based models \cite{wang2026spatio}, and Transformer-based models \cite{piantadosi2024photovoltaic}. For example, Liu \textit{et al.} proposed a PBiLSTM-CNN model that combines multimodal decomposition with parallel bidirectional long short-term memory and convolutional neural network to extract key meteorological and temporal features, thus improving the accuracy of short-term PV power forecasting \cite{liu2024short}. Kim \textit{et al.} proposed PVTransNet, a Transformer-based model that integrates historical PV power, weather observations, weather forecasts, and solar geometry data for multi-step day-ahead forecasting, thus improving the accuracy of next-day PV power prediction \cite{kim2024multi}. Beyond short-term and day-ahead forecasting, patch-based long-sequence architectures have also been explored for mid-term PV forecasting; for example, Tian and Liang proposed PVMTF, an end-to-end framework that combines patch-based temporal modeling with information-fusion coding to enhance mid-term PV power prediction \cite{TIAN2025126263}.

Recent studies have increasingly focused on incorporating physical constraints into PV power forecasting, as forecasting accuracy is closely related to whether the model can respect the inherent characteristics of solar generation, such as non-generation at night, capacity limits, and ramping behavior under irradiance changes. For example, Liu \textit{et al.} proposed a dual-physics-driven framework that integrates physics-consistent data reconstruction with a boundary-constrained DLSTM predictor, thereby improving defective PV data correction and robust rooftop PV forecasting \cite{liu2026dual}. In addition, the emergence of large language models and time-series foundation models has introduced a new modeling paradigm for PV forecasting, in which general temporal knowledge learned from large-scale data can be adapted to energy time-series prediction. For example, Gao \textit{et al.} proposed an LLM-based reprogramming framework that transfers prior knowledge from a frozen pre-trained LLM to few-shot day-ahead solar-irradiation forecasting \cite{gao2026reprogramming}. He \textit{et al.} proposed a prompt-learning and two-stage fine-tuned LLM framework that transforms PV forecasting into a text-generation task and aligns textual prompts with numerical features through cross-attention and DQ-LoRA \cite{he2026photovoltaic}. Furthermore, since meteorological variables are strongly correlated with PV generation, exploring effective methods for weather information fusion has become a central topic in recent PV forecasting research. For example, Dai \textit{et al.} proposed a PV forecasting method based on weather-type credibility prediction and multi-model dynamic combination, which estimates the reliability of day-ahead weather-type forecasts and adaptively selects forecasting models \cite{dai2025short}.

Although these studies have advanced PV forecasting by incorporating physical constraints, adapting foundation-model architectures, and exploiting meteorological information, several challenges remain. The research gap can be summarized as follows: (1) Many existing methods still treat physical knowledge as an auxiliary constraint rather than integrating it throughout the forecasting process, which may limit their ability to model PV generation under different operating regimes. (2) Despite the growing interest in LLM-based forecasting, how to effectively adapt general temporal priors to PV-specific dynamics remains insufficiently explored. (3) Meteorological information is often introduced as input features, but its interaction with temporal patterns and changing PV operating states is not always explicitly modeled. (4) Most forecasting models are still optimized with generic error objectives, making it difficult to balance prediction accuracy across normal, extreme, and transition periods. These gaps indicate that PV forecasting research still has considerable room for improvement.

\subsection{Contributions}

In this paper, we propose the PARA-PV model, a physics-aware retrieval-augmented framework for PV power forecasting. Specifically, PARA-PV first encodes multivariate PV observations into patch-level temporal representations and retrieves physically consistent historical patterns from a memory bank. It then calibrates the prediction using a frozen Chronos temporal prior, corrects regime-dependent distribution shifts through a physics-aware correction module, and trains the model with a customized loss function that assigns explicit importance to peak, ramping, night-time, and regular generation periods. The major contributions are summarized as follows:

\begin{itemize}
    \item \textbf{\textit{A physics-aware retrieval-augmented learner module for PV forecasting:}}
    We develop PA-RAL, a retrieval-augmented forecasting module that incorporates historical PV patterns under physically comparable operating conditions. PA-RAL maintains a patch memory bank that stores two complementary forms of historical information: latent patch representations for feature-level memory enhancement and analog power trajectories for trajectory-level forecasting guidance. During the forecasting process, the current sequence is encoded into patch-level representations and retrieves top-ranked historical patches using a similarity score that jointly considers temporal shape, power-level consistency, PV-state agreement, and intra-day proximity. The retrieved patch representations are transformed into a local memory and fused with the global temporal memory extracted from the current input through a gate mechanism. In parallel, the corresponding retrieved analog trajectories are aggregated according to retrieval confidence and aligned to the latest observed power value, so that their evolution patterns can serve as a trajectory prior without directly copying mismatched absolute magnitudes. A learnable gate then determines the contribution of this before the output.

    \item \textbf{\textit{A Chronos-guided temporal prior calibration strategy to enhance PV prediction:}}
    We introduce this module to exploit the general temporal modeling capability of a time-series foundation model while adapting its output to the physical characteristics of PV generation. Specifically, the normalized historical PV sequence is fed into a frozen Chronos model, which produces a future temporal prior for the target power trajectory. This prior is not directly used as the final forecast because a foundation model trained on broad time-series data may not fully capture local PV operating constraints. Instead, our model combines the Chronos prior with the PA-RAL base prediction, their residual difference, recent power observations, local variability, and PV-state metadata as the input to a lightweight residual adapter. The adapter estimates a correction term, which is added to the PA-RAL output to form the calibrated prediction.

    \item \textbf{\textit{A physics-aware distribution shift correction module for reducing prediction bias:}}
    PV power generation is highly sensitive to weather variability, day-night transitions, and operating regimes, which can lead to conditional distribution shifts between historical training samples and future forecasting scenarios. To address this problem, we design PA-DSC as a post-prediction correction module that adjusts the prediction. Specifically, PA-DSC constructs four input channels: the historical power sequence, a TOPSIS-based weather score, timestamp information, and a day/night indicator. These physical and temporal conditions are first projected into a hidden representation and then processed by a compact dilated temporal convolutional network to capture local and multi-scale correction patterns. Finally, the module outputs three correction factors, namely a mean shift, a scale adjustment, and a correction gate, which are applied to the preliminary target-power prediction in a gated manner.

    \item \textbf{\textit{A physics-constrained loss function for imbalanced-sample optimization:}}
    We formulate a data-driven segment-balanced loss function to guide model training toward PV-relevant operating regimes. Standard forecasting losses assign the same importance to all time steps, which may weaken the learning signal for sparse but operationally important cases such as peak generation and rapid ramping. To address this limitation, the proposed objective restores the target sequence to the physical power scale, partitions samples into peak, ramping, night-time, and regular regimes, and adaptively balances their loss contributions. This design encourages the model to improve overall forecasting accuracy while maintaining sensitivity to physically critical PV generation patterns.
\end{itemize}

The rest of the paper proceeds as follows: Section 2 introduces the problem formulation. Section 3 presents the proposed PARA-PV methodology in detail, covering the overall framework, the physics-aware retrieval-augmented learner, Chronos-guided temporal prior calibration, physics-aware distribution shift correction, and the physics-constrained learning loss function. Section 4 reports the experimental verification, including the experimental settings, comparative studies for point and probabilistic forecasting, ablation analysis, computational complexity evaluation, and sensitivity study. Finally, Section 5 concludes this paper.

\section{Problem Formulation}

In this section, we first define the multivariate PV observation sequence and the historical input window. Then, we formulate deterministic point forecasting and probabilistic forecasting.

\textit{\textbf{Definition 1}} PV data: We define the multivariate PV data as $\mathbf{x}_t = [\mathbf{z}_t, y_t] \in \mathbb{R}^{C}$, where $y_t \in \mathbb{R}$ is the PV power output at time step $t$, and $\mathbf{z}_t \in \mathbb{R}^{C-1}$ denotes the associated covariates, such as solar irradiance, temperature, humidity, atmospheric pressure, and temporal markers. Given a look-back window of length $L$, the historical input is defined as $\mathbf{X}_{t-L+1:t}=[\mathbf{x}_{t-L+1}, \mathbf{x}_{t-L+2}, \ldots, \mathbf{x}_{t}]\in \mathbb{R}^{L \times C}$. The corresponding future PV power trajectory over a prediction horizon $H$ is $\mathbf{Y}_{t+1:t+H}=[y_{t+1}, y_{t+2}, \ldots, y_{t+H}]\in \mathbb{R}^{H}$.

\textit{\textbf{Definition 2}} PV point forecasting problem: Given a historical observation window $\mathbf{X}_{t-L+1:t}$, the objective is to learn a parameterized mapping $f_{\theta}: \mathbb{R}^{L \times C} \rightarrow \mathbb{R}^{H}$ that captures temporal dependencies and produces a deterministic estimate of the future PV power trajectory:
\begin{equation}
    \hat{\mathbf{Y}}_{t+1:t+H}
    =
    f_{\theta}(\mathbf{X}_{t-L+1:t}).
\end{equation}
The model parameters $\theta$ are optimized by minimizing the forecasting error over all training windows:
\begin{equation}
    \theta^{*}
    =
    \arg\min_{\theta}
    \frac{1}{N}
    \sum_{i=1}^{N}
    \mathcal{L}_{\mathrm{point}}
    \left(
        \hat{\mathbf{Y}}^{(i)}_{t+1:t+H},
        \mathbf{Y}^{(i)}_{t+1:t+H}
    \right),
\end{equation}
where $N$ is the number of training samples and $\mathcal{L}_{\mathrm{point}}(\cdot)$ denotes a point-wise forecasting loss.

\textit{\textbf{Definition 3}} PV probabilistic forecasting problem: Given the historical window $\mathbf{X}_{t-L+1:t}$, probabilistic forecasting aims to estimate the conditional distribution $p_{\theta}(\mathbf{Y}_{t+1:t+H}\mid\mathbf{X}_{t-L+1:t})$, thus quantifying the uncertainty of future PV generation. In this work, the distribution is represented by predictive quantiles. For each quantile level $\tau \in \mathcal{T}$, the model outputs:
\begin{equation}
    \hat{\mathbf{Q}}_{\tau,t+1:t+H}
    =
    q_{\theta}^{\tau}
    \left(
        \mathbf{X}_{t-L+1:t}
    \right)
    \in \mathbb{R}^{H},
\end{equation}
where $\hat{\mathbf{Q}}_{\tau,t+1:t+H}$ denotes the conditional $\tau$-quantile of the future PV power. The model parameters are learned by minimizing the average pinball loss over all training samples and quantile levels:
\begin{equation}
    \theta^{*}
    =
    \arg\min_{\theta}
    \frac{1}{N|\mathcal{T}|}
    \sum_{i=1}^{N}
    \sum_{\tau \in \mathcal{T}}
    \rho_{\tau}
    \left(
        \mathbf{Y}^{(i)}_{t+1:t+H}
        -
        \hat{\mathbf{Q}}^{(i)}_{\tau,t+1:t+H}
    \right),
\end{equation}
where $\rho_{\tau}(u)=\max(\tau u,(\tau-1)u)$ is the pinball loss.

\section{Methodology}

\subsection{Overall framework of PARA-PV}

\begin{figure}[!h]
    \centering
    \includegraphics[width=1.0\textwidth]{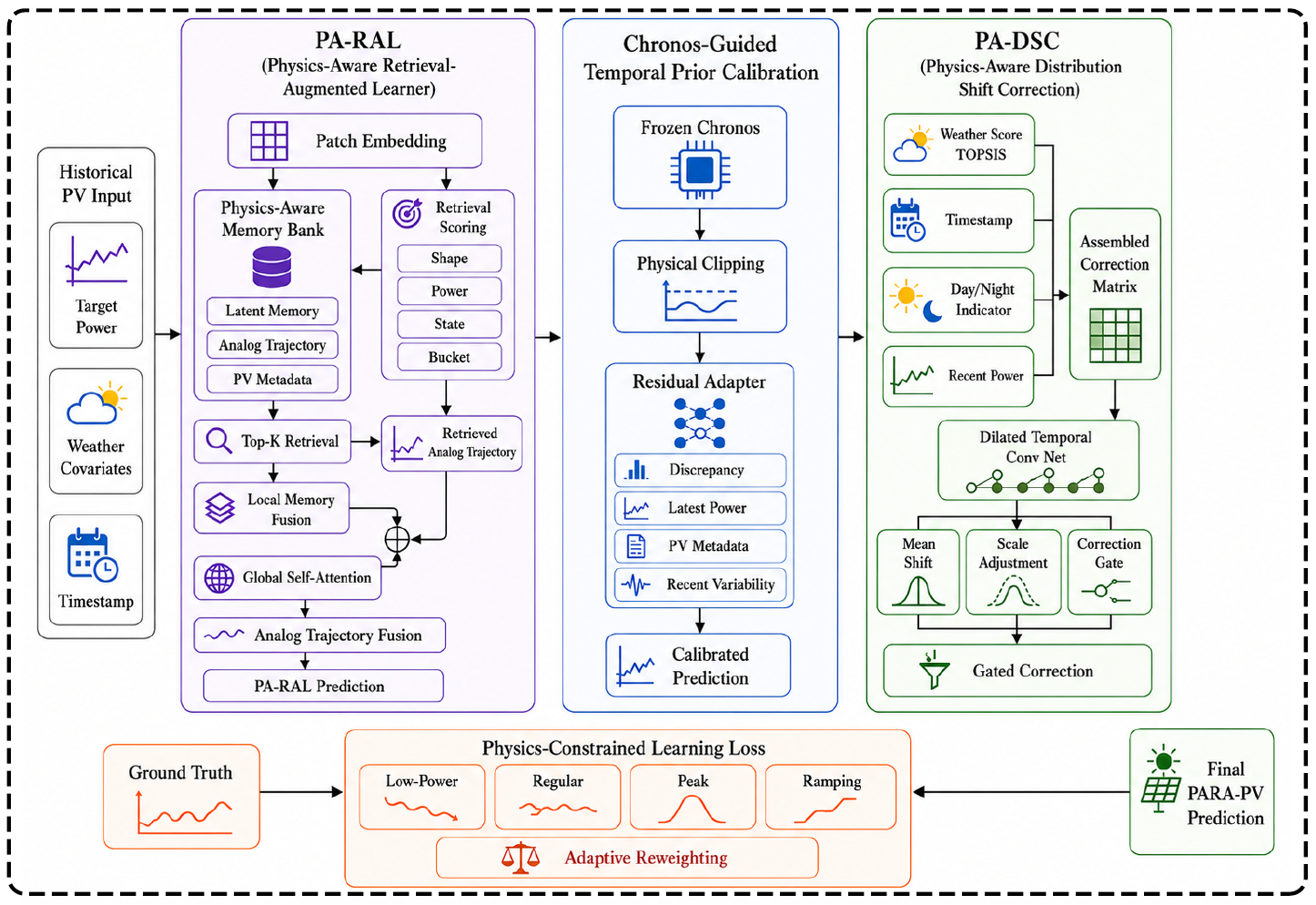}
    \vspace{-1.4\baselineskip}
    \caption{Architecture of the overall framework.}
    \label{fig:framework}
\end{figure}

PARA-PV is designed as a physics-aware forecasting framework that integrates historical analog retrieval, foundation-model temporal priors, distribution correction, and a physically-guided loss function for PV power forecasting. First, the physics-aware retrieval-augmented learner (PA-RAL) retrieves physically consistent historical patches from a memory bank and fuses their latent representations and analog trajectories to produce the base forecast. Second, a frozen Chronos time-series foundation model provides a general temporal prior, which is calibrated by a lightweight residual adapter and fused with the PA-RAL prediction to improve PV-specific temporal modeling. Third, the physics-aware distribution shift correction module (PA-DSC) uses recent power observations, weather-derived scores, timestamp information, and day/night indicators to correct regime-dependent prediction bias after forecasting. Finally, a physics-constrained learning loss partitions PV samples into operating regimes and adaptively reweights forecasting errors, encouraging the model to learn both dominant generation patterns and important low-power, peak, and ramping states. The detailed architecture is shown in \textbf{Fig.~\ref{fig:framework}}.

\subsection{Physics-aware retrieval-augmented Learner (PA-RAL)}

\begin{figure}[!h]
    \centering
    \includegraphics[width=1.0\textwidth]{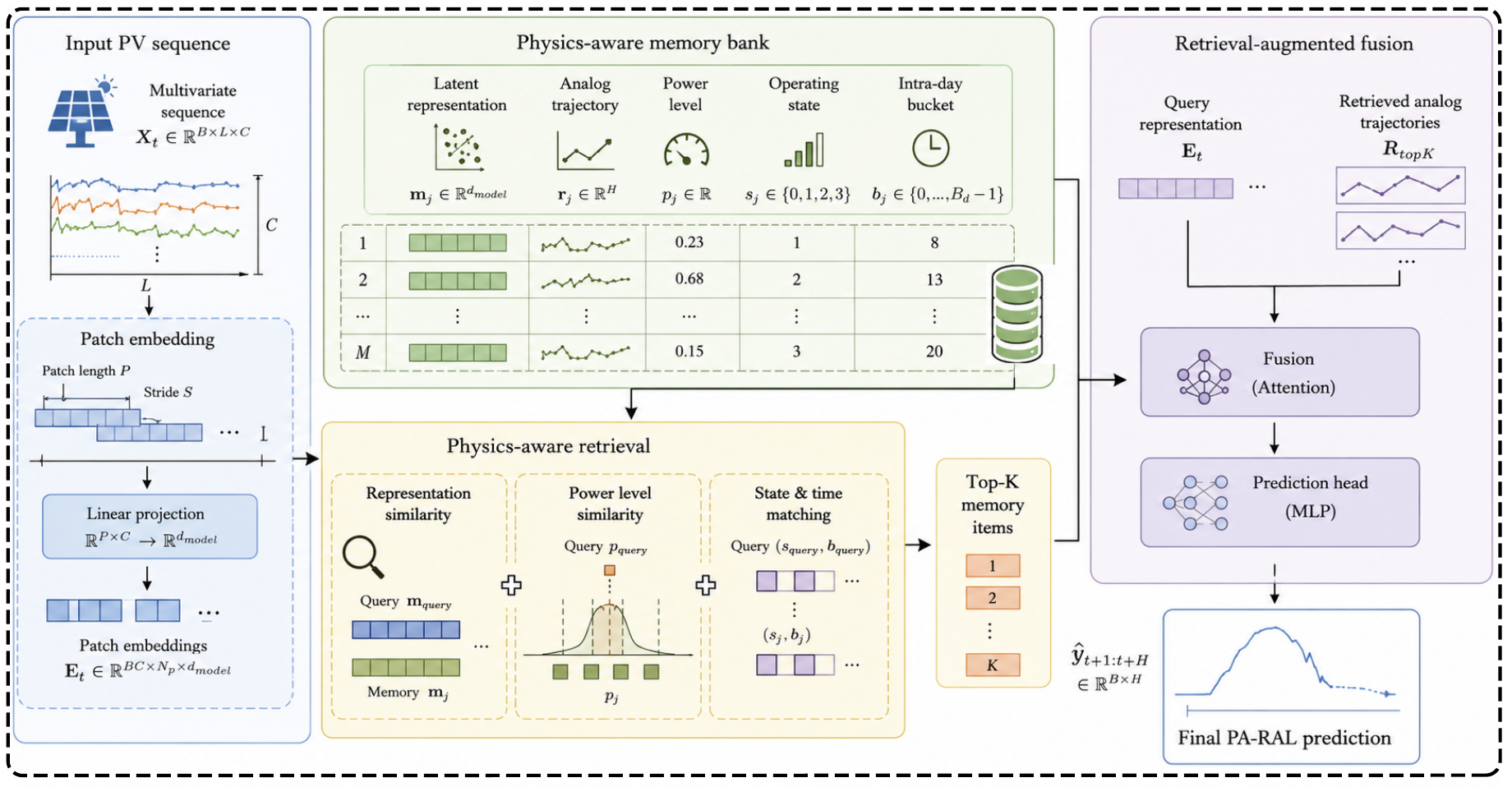}
    \vspace{-1.4\baselineskip}
    \caption{Architecture of the physics-aware retrieval-augmented learner module.}
    \label{fig:pa_ral_module}
\end{figure}

PV power forecasting remains difficult because future generation cannot be inferred from temporal patterns alone. It depends on the interaction between historical dynamics, meteorological conditions, diurnal cycles, and the physical operating regime of the PV system. Two PV sequences may show similar short-term variations but correspond to different generation states, such as low output, rapid ramping, or peak production. Conversely, physically comparable sequences may contain useful information even when their raw temporal shapes differ. Most temporal models encode the current input window as a latent representation and learn a direct mapping to future power, which makes it difficult to reuse historical PV behaviors observed under similar physical conditions explicitly. Retrieval-based methods can introduce historical context; however, retrieval based solely on feature similarity may select samples that are close in representation space while differing in power level, operating state, or intra-day position. We design the Physics-Aware Retrieval-Augmented Learner (PA-RAL) to retrieve and integrate historical PV patterns under physically comparable conditions. PA-RAL combines patch-level temporal retrieval with PV-specific metadata, including power level, operating state, and intra-day bucket, allowing the model to build a physically meaningful memory rather than a purely data-driven nearest-neighbor set. As a result, the forecasting process can use both the learned representation of the current input and historical generation patterns associated with similar PV regimes. The detailed architecture is illustrated in \textbf{Fig.~\ref{fig:pa_ral_module}}.

\subsubsection{Patch representation and physics-aware memory bank}

\textbf{Patch embedding:}
Given a normalized multivariate PV input sequence
$\mathbf{X}_{t} \in \mathbb{R}^{B \times L \times C}$, where $B$, $L$, and $C$ denote the batch size, sequence length, and number of variables, respectively. PA-RAL first divides each channel-wise time series into overlapping patches of length $P$ with stride $S$. Each patch is linearly projected into a $d_{\mathrm{model}}$-dimensional latent space, producing patch embeddings
\begin{equation}
    \mathbf{E}_{t}
    =
    \mathrm{PatchEmbed}(\mathbf{X}_{t})
    \in
    \mathbb{R}^{B C \times N_p \times d_{\mathrm{model}}},
\end{equation}
where $N_p=\left\lfloor\frac{L + P_{\mathrm{ad}} - P}{S}\right\rfloor+ 1$ is the number of patches and $P_{\mathrm{ad}}$ denotes the padding length used before patch extraction. In this form, each variable channel is represented as a sequence of local temporal tokens, enabling retrieval to be performed on short-term temporal patterns rather than on the raw input sequence.

\textbf{Physics-aware memory bank:}
It is used to store historical PV patterns in both latent and physical forms. The complete memory bank is defined as:
\begin{equation}
    \mathcal{M}
    =
    \left\{
    \mathcal{M}_j
    \right\}_{j=1}^{M}
    =
    \left\{
    \left(
        \mathbf{m}_j,
        \mathbf{r}_j,
        p_j,
        s_j,
        b_j
    \right)
    \right\}_{j=1}^{M},
\end{equation}
where $M$ is the maximum memory size. Each memory item contains a compact latent representation, an analog trajectory, and three PV-related metadata variables. This design allows the retrieval process to consider not only representation similarity, but also the physical operating condition under which each historical pattern was observed.

The first component, $\mathbf{m}_j\in\mathbb{R}^{d_{\mathrm{model}}}$, is the stored latent representation used for feature-level retrieval. Let $(t_j,c_j)$ denote the forecasting origin and variable channel associated with the $j$-th memory item. We define a shared mean-pooling operator for a channel-wise patch sequence as:
\begin{equation}
\mathcal{P}\left(\mathbf{E}_{t_j,c_j}\right)
=
\frac{1}{N_p}
\sum_{n=1}^{N_p}
\mathbf{E}_{t_j,c_j,n}.
\label{7}
\end{equation}

The latent key stored in the $j$-th memory item is then defined as:
\begin{equation}
\mathbf{m}_j
=
\mathcal{P}\left(\mathbf{E}_{t_j,c_j}\right)
\in\mathbb{R}^{d_{\mathrm{model}}}.
\end{equation}

The second component, $\mathbf{r}_j \in \mathbb{R}^{H}$, denotes the historical analog trajectory, where $H$ is the prediction horizon. The $\mathbf{r}_j$ is extracted from the most recent $H$ time steps of the normalized historical input sequence associated with the memory item. Hence, it does not contain future-label information; instead, it preserves the recent evolution shape of a historical PV pattern and is later used as a trajectory-level prior.

The remaining components describe the physical context associated with each memory item. Specifically, $p_j \in \mathbb{R}$ denotes the average target-power level of the corresponding historical window measured in the physical power scale, $s_j$ denotes a categorical PV operating state, and $b_j$ denotes the intra-day temporal bucket. The PV state is defined as $s_j \in \{0,1,2,3\}$, corresponding to low-power, regular generation, peak-generation, and ramping states, respectively. The temporal bucket $b_j \in \{0,1,\ldots, B_d-1\}$ records the time-of-day position of the memory item, where $B_d$ is the number of daily buckets; in our implementation, $B_d=24$.

These metadata are computed from the historical PV window before the normalization. For a window associated with the $j$-th memory item, let $\mathbf{y}_j=[y_{j,1},\ldots,y_{j,L}]$ denote the target-power sequence. The power level is defined as:
\begin{equation}
    p_j = \frac{1}{L}\sum_{\ell=1}^{L} y_{j,\ell}.
\end{equation}

To assign the operating state, we compute the maximum power and ramping strength of the window:
\begin{equation}
    y_j^{\max} = \max_{\ell} y_{j,\ell}, \qquad
    \Delta_j = \max_{\ell>1} |y_{j,\ell}-y_{j,\ell-1}|.
\end{equation}

Given data-driven thresholds for low-power, peak-generation, and ramping regimes, denoted by $\tau_{\mathrm{low}}$, $\tau_{\mathrm{peak}}$, and $\tau_{\mathrm{ramp}}$, respectively, which are estimated from the training split. Specifically, $\tau_{\mathrm{low}}$ is set to the 1st percentile of positive training power values, $\tau_{\mathrm{peak}}$ is set to the 90th percentile of non-low-power training samples, and $\tau_{\mathrm{ramp}}$ is set to the 80th percentile of non-zero absolute power differences between adjacent training time steps. The state label is assigned as:
\begin{equation}
s_j =
\begin{cases}
0, & p_j \leq \tau_{\mathrm{low}},\\
2, & y_j^{\max} > \tau_{\mathrm{peak}},\\
3, & \Delta_j \geq \tau_{\mathrm{ramp}},\ p_j > \tau_{\mathrm{low}},\
     y_j^{\max} \leq \tau_{\mathrm{peak}},\\
1, & \mathrm{otherwise}.
\end{cases}
\end{equation}

The intra-day bucket $b_j$ is obtained by discretizing the hour of the last time step in the window into $B_d$ circular daily intervals. Hence, $\mathbf{m}_j$ supports retrieval in the latent representation space, $\mathbf{r}_j$ provides analogical trajectory guidance, and $(p_j,s_j,b_j)$ constrains retrieval toward historical samples with comparable PV operating conditions and intra-day positions.

\subsubsection{Physics-aware retrieval mechanism}

Given the physics-aware memory bank, PA-RAL retrieves historical samples by jointly considering latent temporal similarity and PV operating-context consistency. For the current input at forecasting origin $t$, the patch-token sequence of the $c$-th variable channel is denoted by $\mathbf{E}_{t,c}$. Using the same pooling operator defined in Eq.~(\ref{7}), we construct the current query as:
\begin{equation}
\mathbf{q}_{t,c}
=
\mathcal{P}\left(\mathbf{E}_{t,c}\right)
\in\mathbb{R}^{d_{\mathrm{model}}}.
\end{equation}

For each memory item $\mathcal{M}_j=(\mathbf{m}_j,\mathbf{r}_j,p_j,s_j,b_j)$, PA-RAL computes four similarity terms. The first term measures temporal-shape similarity in the latent representation space by cosine similarity:
\begin{equation}
    S_{\mathrm{shape}}(\mathbf{q}_{t,c},\mathbf{m}_j)
    =
    \frac{
        \mathbf{q}_{t,c}^{\top}\mathbf{m}_j
    }{
        \|\mathbf{q}_{t,c}\|_2 \|\mathbf{m}_j\|_2
    }.
\end{equation}

The second term measures the consistency of power levels between the current window and the historical memory item:
\begin{equation}
    S_{\mathrm{power}}(p_t,p_j)
    =
    \frac{1}
    {1 + |p_t-p_j|/\sigma_p},
\end{equation}
where $\sigma_p$ is a scale factor used to normalize power-level differences. The third term encourages retrieval from the same PV operating state:
\begin{equation}
    S_{\mathrm{state}}(s_t,s_j)
    =
    \mathbb{I}(s_t=s_j),
\end{equation}
where $\mathbb{I}(\cdot)$ is the indicator function. The fourth term measures the proximity of intra-day positions. Since the time of day is periodic, we first compute the circular bucket distance:
\begin{equation}
    d_b(b_t,b_j)
    =
    \min\left(
        |b_t-b_j|,
        B_d-|b_t-b_j|
    \right),
\end{equation}
and then define the bucket similarity as:
\begin{equation}
    S_{\mathrm{bucket}}(b_t,b_j)
    =
    \max\left(
        0,
        1-\frac{d_b(b_t,b_j)}{B_d/2}
    \right).
\end{equation}

The final retrieval score is obtained by a learnable weighted combination of the four similarity terms:
\begin{equation}
\begin{aligned}
    S_{t,c,j}
    ={}&
    \omega_1 S_{\mathrm{shape}}(\mathbf{q}_{t,c},\mathbf{m}_j)
    + \omega_2 S_{\mathrm{power}}(p_t,p_j) \\
    &+
    \omega_3 S_{\mathrm{state}}(s_t,s_j)
    + \omega_4 S_{\mathrm{bucket}}(b_t,b_j),
\end{aligned}
\end{equation}
where the weights are normalized by:
\begin{equation}
    \boldsymbol{\omega}
    =
    \mathrm{softmax}(\boldsymbol{\eta}),
    \qquad
    \boldsymbol{\omega}
    =
    [\omega_1,\omega_2,\omega_3,\omega_4],
\end{equation}
where, $\boldsymbol{\eta}$ is a set of learnable retrieval logits. This design allows PA-RAL to adaptively determine the relative importance of temporal shape, power level, PV state, and intra-day position during training. Based on the retrieval score, PA-RAL selects the top-$K$ memory items:
\begin{equation}
    \mathcal{N}_{K}(t,c)
    =
    \operatorname{TopK}_{j}
    \left(
        S_{t,c,j}
    \right),
\end{equation}
and converts their scores into normalized retrieval weights:
\begin{equation}
    \alpha_{t,c,j}
    =
    \frac{
        \exp(S_{t,c,j})
    }{
        \sum_{k\in\mathcal{N}_{K}(t,c)}
        \exp(S_{t,c,k})
    },
    \qquad
    j\in\mathcal{N}_{K}(t,c).
\end{equation}

The retrieved latent representations $\{\mathbf{m}_j\}_{j\in\mathcal{N}_{K}(t,c)}$ are used to construct a feature-level local memory. Therefore, PA-RAL retrieves historical patterns that are not only close in learned temporal representation but also consistent with the physical generation regime and intra-day context of the current window.

\subsubsection{Local and global memory fusion}

After retrieving physically consistent historical samples, PA-RAL fuses the retrieved external memory with the temporal dependencies extracted from the current input window. For the query associated with the $c$-th variable channel at forecasting origin $t$, let $\mathcal{N}_{K}(t,c)$ denote the set of top-$K$ retrieved memory indices. The corresponding latent representations $\{\mathbf{m}_j\}_{j\in\mathcal{N}_{K}(t,c)}$ are first transformed by a local memory network $f_{\mathrm{loc}}(\cdot)$ and then averaged to obtain a retrieved local context:
\begin{equation}
    \mathbf{h}^{\mathrm{ret}}_{t,c}
    =
    \frac{1}{K}
    \sum_{j\in\mathcal{N}_{K}(t,c)}
    f_{\mathrm{loc}}(\mathbf{m}_j)
    \in
    \mathbb{R}^{d_{\mathrm{model}}}.
\end{equation}

This retrieved context is injected into the current patch-token sequence through
a residual connection:
\begin{equation}
    \mathbf{H}^{\mathrm{loc}}_{t,c,n}
    =
    \mathbf{E}_{t,c,n}
    +
    \mathbf{h}^{\mathrm{ret}}_{t,c},
    \qquad
    n=1,\ldots,N_p .
\end{equation}

Thus, each current patch token is enhanced by historical contexts retrieved from physically comparable PV regimes, while the original temporal representation is preserved.

In parallel, PA-RAL computes a global memory from the current patch sequence using multi-head self-attention. Let $\mathbf{E}_{t,c}=[\mathbf{E}_{t,c,1},\ldots,\mathbf{E}_{t,c,N_p}]$ denote the patch-token sequence of one channel. The global dependency representation is computed as:
\begin{equation}
    \mathbf{Z}_{t,c}
    =
    \mathrm{MHA}
    \left(
        \mathbf{E}_{t,c},
        \mathbf{E}_{t,c},
        \mathbf{E}_{t,c}
    \right)
    \in
    \mathbb{R}^{N_p \times d_{\mathrm{model}}},
\end{equation}
where $\mathrm{MHA}(\cdot)$ denotes multi-head self-attention. This operation captures the internal temporal dependencies among patches within the current input window. The global context is summarized as:
\begin{equation}
    \mathbf{h}^{\mathrm{glob}}_{t,c}
    =
    \frac{1}{N_p}
    \sum_{n=1}^{N_p}
    \mathbf{Z}_{t,c,n}
    \in
    \mathbb{R}^{d_{\mathrm{model}}}.
\end{equation}

The local and global memories are finally combined to form the memory-enhanced representation:
\begin{equation}
    \mathbf{H}^{\mathrm{mem}}_{t,c,n}
    =
    \mathbf{H}^{\mathrm{loc}}_{t,c,n}
    +
    \mathbf{h}^{\mathrm{glob}}_{t,c},
\end{equation}
where, $\mathbf{H}^{\mathrm{loc}}_{t,c,n}$ carries historical information
retrieved from physically similar PV samples, whereas $\mathbf{h}^{\mathrm{glob}}_{t,c}$ encodes the temporal structure of the current input sequence. The fused representation is flattened and mapped to the prediction horizon by a forecasting head:
\begin{equation}
    \hat{\mathbf{y}}^{\mathrm{mem}}_{t,c}
    =
    f_{\mathrm{pred}}
    \left(
        \mathrm{Flatten}
        \left(
            \mathbf{H}^{\mathrm{mem}}_{t,c}
        \right)
    \right)
    \in
    \mathbb{R}^{H},
\end{equation}
where $H$ is the prediction horizon. The resulting
$\hat{\mathbf{y}}^{\mathrm{mem}}_{t,c}$ serves as the memory-based base
prediction of PA-RAL, which is further refined by the trajectory-level analog prior in the subsequent fusion stage.

\subsubsection{Analog trajectory enhanced fusion}

The retrieved latent patch representations enhance the representation of the current input, but they do not explicitly describe how PV power evolved in the corresponding historical windows. To further exploit this trajectory-level information, PA-RAL uses the analog trajectories stored in the same retrieved memory items as an additional forecasting prior. 

For the query associated with channel $c$ at forecasting origin $t$, the normalized analog prior is computed as:
\begin{equation}
\bar{\mathbf{a}}_{t,c}
=
\sum_{j\in\mathcal{N}_K(t,c)}
\alpha_{t,c,j}\bar{\mathbf{r}}_j,
\end{equation}
where $\bar{\mathbf{r}}_j\in\mathbb{R}^{H}$ is the normalized trajectory attached to the $j$-th retrieved memory item. The analog prior is then aligned with the latest normalized observation of the query channel:
\begin{equation}
\tilde{\mathbf{a}}_{t,c}
=
\bar{x}_{t,c}\mathbf{1}_{H}
+
\left(
\bar{\mathbf{a}}_{t,c}
-
\bar{a}_{t,c,1}\mathbf{1}_{H}
\right).
\label{29}
\end{equation}

All terms in Eq.~(\ref{29}) are dimensionless quantities in the window-normalized space. After prediction and analog fusion, the output is transformed back to the original scale using the mean and standard deviation of the corresponding query channel.

We further estimate the reliability of the retrieved analog prior. When the retrieval weights are close to a uniform distribution, no single historical trajectory is strongly supported by the retrieval mechanism; when one retrieved item dominates, the analog prior is more reliable. For $K>1$, the score is defined as:
\begin{equation}
    \rho_{t,c}
    =
    \frac{
        \max_{j\in\mathcal{N}_{K}(t,c)}\alpha_{t,c,j}
        -
        1/K
    }{
        1-1/K
    },
    \qquad
    \rho_{t,c}\in[0,1].
\end{equation}
This score is used to modulate the contribution of the analog prior during fusion.

Let $\hat{\mathbf{y}}^{\mathrm{mem}}_{t,c}\in\mathbb{R}^{H}$ be the
memory-based prediction produced by the local--global memory fusion stage, and let $\mathbf{e}_{t,c}$ denote the PV metadata embedding used for gating. PA-RAL computes a reliability-aware fusion coefficient as:
\begin{equation}
    \lambda_{t,c}
    =
    \rho_{t,c}\,
    g_{\phi}
    \left(
        \left[
            \hat{\mathbf{y}}^{\mathrm{mem}}_{t,c},
            \tilde{\mathbf{a}}_{t,c},
            \rho_{t,c},
            \mathbf{e}_{t,c}
        \right]
    \right),
\end{equation}
where $g_{\phi}(\cdot)$ is a learnable gating network and
$\lambda_{t,c}\in[0,1]$. The final PA-RAL prediction is then obtained by a
convex interpolation between the memory-based prediction and the aligned analog
prior:
\begin{equation}
    \hat{\mathbf{y}}^{\mathrm{PA\mbox{-}RAL}}_{t,c}
    =
    (1-\lambda_{t,c})
    \hat{\mathbf{y}}^{\mathrm{mem}}_{t,c}
    +
    \lambda_{t,c}
    \tilde{\mathbf{a}}_{t,c}.
\end{equation}

Through this, PA-RAL injects trajectory-level historical guidance only when the retrieved analogs are reliable with the current forecasting condition. The model can benefit from historical PV evolution patterns without directly copying trajectories whose absolute magnitudes may be mismatched.

\subsection{Chronos-guided temporal prior calibration}

Although PA-RAL brings physically consistent historical retrieval into PV forecasting, its predictions still mainly depend on patterns in the available PV data and the retrieved local memory. This works well when similar operating regimes are well represented in the memory bank, but it can be less reliable when the current forecasting window contains temporal dynamics that are rarely seen in local historical samples. Time-series foundation models offer a natural complement because they learn general temporal structures from large and diverse time-series corpora, providing a broader prior beyond local PV memory. Directly using such a model as a standalone PV forecaster, however, is not suitable, since PV generation is governed by domain-specific physical constraints, such as capacity limits, low nighttime output, and regime-dependent responses to weather conditions. We use Chronos not to replace PA-RAL, but as a frozen temporal prior generator whose output is adapted through PV-specific residual learning. In this way, PARA-PV can draw on the general temporal modeling ability of Chronos while retaining the physically grounded and regime-aware behavior learned by PA-RAL.

Chronos is a time-series foundation model that casts forecasting as sequence modeling \cite{ansari2025chronos2}. Rather than learning a separate forecaster for each task, it learns broad temporal regularities from large time-series corpora by discretizing real-valued observations into tokens and using a pretrained Transformer to model their future evolution. As a result, Chronos can provide transferable forecasting priors for unseen time-series domains, either in a zero-shot setting or as a frozen prior. In this work, we employ the 120M-parameter, encoder-only Chronos-2 model as a frozen temporal prior generator. All Chronos-2 parameters are fixed during training, and only the lightweight residual adapter is optimized for the downstream PV forecasting task. Given the normalized historical PV sequence as $\mathbf{X}_{t-L+1:t} = [\mathbf{z}_{t-L+1:t}, \mathbf{y}_{t-L+1:t}]$, where $\mathbf{y}_{t-L+1:t}$ denotes the historical target-power sequence and $\mathbf{z}_{t-L+1:t}$ denotes the associated covariates. The frozen Chronos model then produces a future quantile forecast over the prediction horizon $H$:

\begin{equation}
    \mathbf{c}_{t}^{\tau}
    =
    \mathcal{F}_{\mathrm{Chronos}}
    \left(
        \mathbf{X}_{t-L+1:t};
        \tau
    \right)
    \in
    \mathbb{R}^{H},
\end{equation}
where $\tau$ is the selected quantile level. In our implementation, we use the median quantile, i.e. $\tau=0.5$, as the temporal prior: $\mathbf{c}_{t} = \mathbf{c}_{t}^{0.5}$.

Before residual calibration, we constrain the Chronos prior using the physical bounds of PV generation. The zero-power level and PV capacity are first mapped to the normalized prediction space, and the Chronos prior is then clipped to this interval. For windows classified as nighttime or low-power states, the prior is further pulled toward the normalized zero-power boundary. This step prevents the general-purpose temporal prior from producing physically infeasible PV estimates.

To calibrate the Chronos prior, we construct a context-aware input for a lightweight residual adapter. Let $\hat{\mathbf{y}}^{\mathrm{ral}}_{t}\in\mathbb{R}^{H}$ denote the PA-RAL base prediction for the target-power channel, and let $\bar{\mathbf{c}}_{t}\in\mathbb{R}^{H}$ denote the physically clipped Chronos prior over the same horizon. We first compute their discrepancy as:
\begin{equation}
    \mathbf{d}_{t}
    =
    \bar{\mathbf{c}}_{t}
    -
    \hat{\mathbf{y}}^{\mathrm{ral}}_{t},
\end{equation}
which represents the correction direction suggested by the frozen temporal foundation model relative to the PV-aware PA-RAL prediction. 

In addition, we include the latest normalized target-power value expanded to the prediction horizon, which is defined as $\mathbf{l}_{t} = x^{\mathrm{pow}}_{t}\mathbf{1}_{H} \in \mathbb{R}^{H}$. The PV operating metadata inherited from PA-RAL is represented as: $\mathbf{e}^{\mathrm{pv}}_{t} = \left[ \mathrm{OneHot}(s_t), \tilde{p}_t, \tilde{b}_t \right]$, where $\mathrm{OneHot}(s_t)$ encodes the PV operating state, $\tilde{p}_t$ is the normalized power level, and $\tilde{b}_t$ is the normalized intra-day bucket. And the recent local variability is defined as $\sigma^{\mathrm{rec}}_{t} = \mathrm{Std} \left( x^{\mathrm{pow}}_{t-L_r+1:t} \right)$, where $L_r$ denotes the length of the recent window used to estimate local power variability. The final adapter input is then defined as:
\begin{equation}
    \mathbf{u}_{t}
    =
    \left[
        \hat{\mathbf{y}}^{\mathrm{ral}}_{t},
        \bar{\mathbf{c}}_{t},
        \mathbf{d}_{t},
        \mathbf{l}_{t},
        \mathbf{e}^{\mathrm{pv}}_{t},
        \sigma^{\mathrm{rec}}_{t}
    \right].
\end{equation}

The residual adapter then maps the calibration input $\mathbf{u}_{t}$ to a PV-specific correction term:
\begin{equation}
    \Delta\mathbf{y}_{t}
    =
    f_{\mathrm{ada}}
    \left(
        \mathbf{u}_{t}
    \right)
    \in
    \mathbb{R}^{H},
\end{equation}
where $f_{\mathrm{ada}}(\cdot)$ is a lightweight multilayer adapter composed of linear projection, layer normalization, nonlinear activation, dropout, and a final linear output layer. Instead of replacing the PA-RAL prediction with the Chronos output, the adapter only learns the residual correction needed to align the PA-RAL forecast with the useful temporal prior provided by Chronos. The calibrated target-power prediction is defined as:
\begin{equation}
    \hat{\mathbf{y}}^{\mathrm{cal}}_{t}
    =
    \hat{\mathbf{y}}^{\mathrm{ral}}_{t}
    +
    \Delta\mathbf{y}_{t}.
\end{equation}

Overall, the Chronos-guided temporal prior calibration module uses Chronos as a frozen source of general temporal knowledge rather than as a standalone PV forecaster. This design preserves the physics-aware retrieval-enhanced prediction of PA-RAL, while allowing PARA-PV to benefit from the broader temporal modeling capability of the foundation model.

\subsection{Physics-aware distribution shift correction (PA-DSC)}

\begin{figure}[!h]
    \centering
    \includegraphics[width=1.0\textwidth]{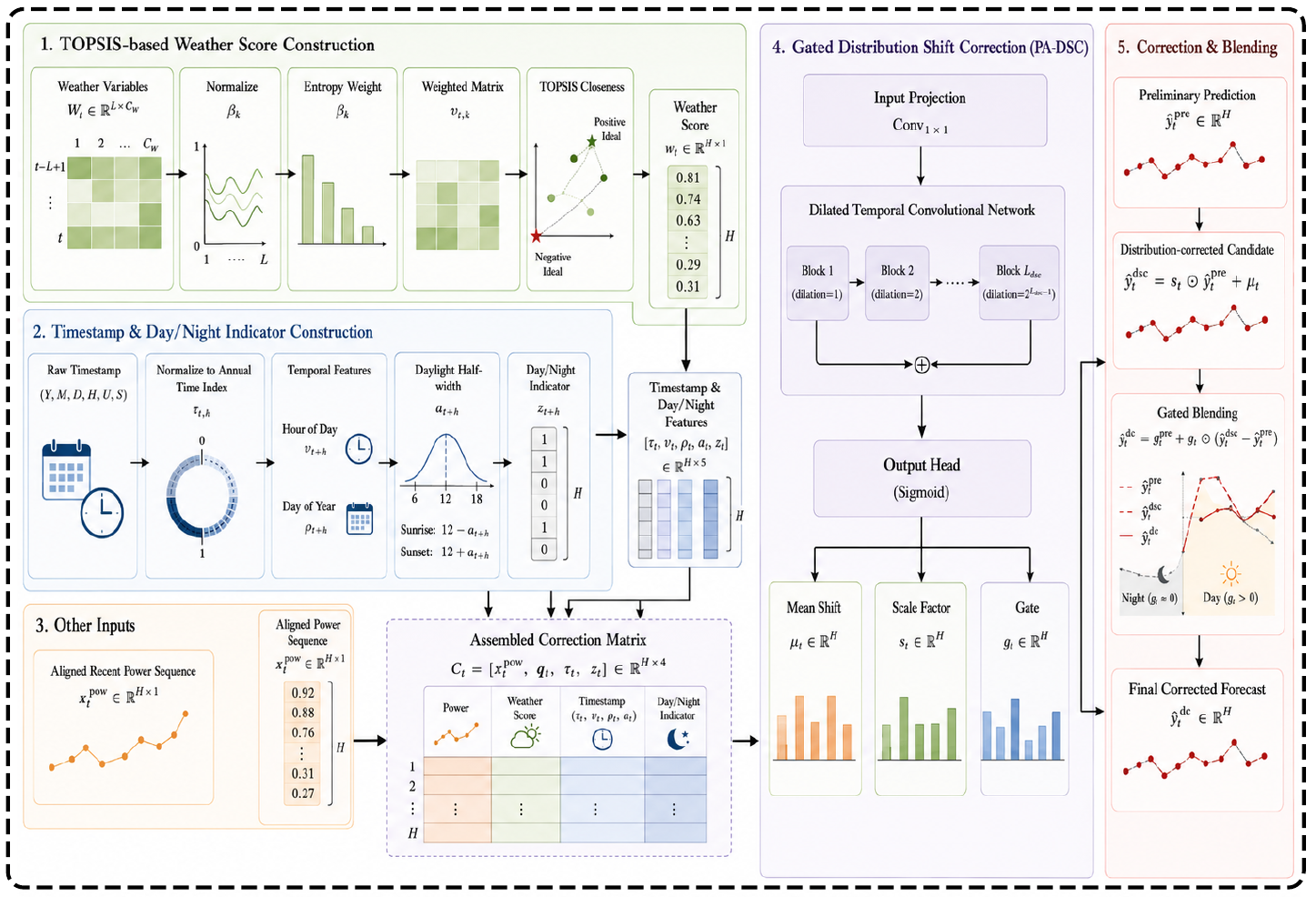}
    \vspace{-1.4\baselineskip}
    \caption{Architecture of the Physics-aware distribution shift correction.}
    \label{fig:pa_dsc_module}
\end{figure}

PV power generation is highly sensitive to weather variability, diurnal transitions, and regime-dependent operating conditions. These factors can create shifts in conditional distributions between historical observations and future forecasting windows. Although PA-RAL and Chronos-guided calibration provide a physics-aware and temporally informed preliminary forecast, the predicted target power may still show systematic bias when weather conditions or day/night regimes change. To this end, we introduce PA-DSC as a post-prediction correction module, which refines the preliminary target-power forecast using weather information, timestamps, and physical day/night cues. This correction reduces regime-dependent bias while maintaining the feasibility of PV generation. \textbf{Fig.~\ref{fig:pa_dsc_module}} shows the architecture of the distribution shift correction module.

\subsubsection{TOPSIS-based weather score construction}

Weather is one of the main sources of distributional variation in PV power. Rather than passing several meteorological variables directly to the correction module, PA-DSC compresses them into a weather score that describes how favorable the current weather state is. Given the normalized historical input sequence, we use the $C_w$ non-target variables as weather covariates. The weather matrix is written as $\mathbf{W}_t = [\mathbf{w}_{t-L+1},\ldots,\mathbf{w}_{t}] \in \mathbb{R}^{L \times C_w}$. Each weather variable is normalized along the temporal dimension:
\begin{equation}
    \tilde{w}_{\ell,k}
    =
    \frac{
        w_{\ell,k}
        -
        \min_{\ell} w_{\ell,k}
    }{
        \max_{\ell} w_{\ell,k}
        -
        \min_{\ell} w_{\ell,k}
        +
        \epsilon
    },
\end{equation}
where $\ell$ indexes time, $k$ indexes the weather variable, and $\epsilon$ is a small constant. The normalized values are then converted into temporal proportions, which are defined as $q_{\ell,k}  = \frac{ \tilde{w}_{\ell,k} }{ \sum_{\ell=1}^{L} \tilde{w}_{\ell,k}  + \epsilon}$. We compute the information entropy of the $k$-th weather variable and its adaptive weight as:
\begin{equation}
    e_k
    =
    -
    \frac{1}{\log L}
    \sum_{\ell=1}^{L}
    q_{\ell,k}\log(q_{\ell,k}+\epsilon),
\end{equation}

\begin{equation}
    \beta_k
    =
    \frac{
        1-e_k
    }{
        \sum_{r=1}^{C_w}(1-e_r)+\epsilon
    }.
\end{equation}

Weather variables with stronger temporal variation and lower entropy are assigned larger weights. The weighted weather representation is given by
\begin{equation}
    v_{\ell,k}
    =
    \beta_k \tilde{w}_{\ell,k}.
\end{equation}

We then measure how close each time step is to the positive and negative ideal weather states using a TOPSIS-style closeness score. The two ideals are defined as $v_k^{+}=\max_{\ell} v_{\ell,k}, \ v_k^{-}=\min_{\ell} v_{\ell,k}$. The distances to the positive and negative ideals are defined as:
\begin{equation}
    D_{\ell}^{+}
    =
    \left(
        \sum_{k=1}^{C_w}
        (v_{\ell,k}-v_k^{+})^2
    \right)^{1/2},
    \qquad
    D_{\ell}^{-}
    =
    \left(
        \sum_{k=1}^{C_w}
        (v_{\ell,k}-v_k^{-})^2
    \right)^{1/2}.
\end{equation}

The weather score at time step $\ell$ is calculated as:
\begin{equation}
    s^{\mathrm{wea}}_{\ell}
    =
    \frac{
        D_{\ell}^{-}
    }{
        D_{\ell}^{+}
        +
        D_{\ell}^{-}
        +
        \epsilon
    }.
\end{equation}

A larger score means that the weather state is closer to the positive ideal and farther from the negative ideal. The weather score sequence is finally aligned with the prediction horizon $H$:
\begin{equation}
    \mathbf{s}^{\mathrm{wea}}_t
    =
    \mathrm{Align}_{H}
    \left(
        [s^{\mathrm{wea}}_{t-L+1},\ldots,s^{\mathrm{wea}}_{t}]
    \right)
    \in
    \mathbb{R}^{H \times 1},
\end{equation}
where $\mathrm{Align}_{H}(\cdot)$ keeps the latest $H$ steps when the sequence is longer than $H$ and pads the sequence with its last value when it is shorter. This TOPSIS-based score gives PA-DSC a compact and interpretable weather signal for correcting weather-induced distribution shifts.

\subsubsection{Timestamp information and day/night indicator construction}

PV generation is also closely tied to temporal position, especially the diurnal cycle. PA-DSC uses two time-related correction signals: a normalized timestamp sequence, which describes the temporal position to describe the variety of seasons, and a day/night indicator, which specifies whether solar generation is physically available at each predicted step.

For each predicted time step $h$, the raw timestamp contains complete calendar information, denoted as $(Y_{t,h}, M_{t,h}, D_{t,h}, H_{t,h}, U_{t,h}, S_{t,h})$, where $Y$, $M$, $D$, $H$, $U$, and $S$ represent year, month, day, hour, minute, and second, respectively. We convert this timestamp into a normalized annual time index. Let $\mathrm{doy}_{t,h}$ be the day-of-year computed from $(Y_{t,h}, M_{t,h}, D_{t,h})$, and let $\mathrm{sod}_{t,h} = 3600H_{t,h} + 60U_{t,h} + S_{t,h}$ denote the seconds elapsed since the beginning of the day. The normalized timestamp is then defined as:
\begin{equation}
    \tau_{t,h}
    =
    \frac{\mathrm{doy}_{t,h}-1}{D_Y}
    +
    \frac{\mathrm{sod}_{t,h}}{D_Y \times 86400},
\end{equation}
where $D_Y$ is the number of days in year $Y_{t,h}$, i.e., $365$ or $366$ for a leap year. The timestamp sequence over the prediction horizon is $\boldsymbol{\tau}_t = [\tau_{t,1},\tau_{t,2},\ldots,\tau_{t,H}] \in \mathbb{R}^{H}$.

Then, the timestamp information is transformed into temporal encoding features. PA-DSC uses the temporal marks of the prediction horizon to derive the hour-of-day $\nu_{t+h}$ and the normalized day-of-year $\rho_{t+h}$. Based on these two quantities, we approximate the seasonal daylight half-width around noon as:
\begin{equation}
    a_{t+h}
    =
    6
    +
    1.5\cos\left(
    2\pi
    \left(
    \rho_{t+h}
    -
    \frac{172}{365}
    \right)
    \right),
\end{equation}
where the constant $172/365$ approximately aligns the maximum daylight duration with the summer solstice. The coefficients $6$ and $1.5$ define the baseline half-daylight duration and its seasonal variation, respectively. Consequently, $12-a_{t+h}$ and $12+a_{t+h}$ approximate the sunrise and sunset times. The day/night indicator is then defined as:
\begin{equation}
    z_{t+h}
    =
    \mathbb{I}
    \left(
    12-a_{t+h}
    \le
    \nu_{t+h}
    \le
    12+a_{t+h}
    \right),
\end{equation}
where $z_{t+h}=1$ denotes a daylight period and $z_{t+h}=0$ denotes a night-time period. The vector $\mathbf{z}_{t}=[z_{t+1},\ldots,z_{t+H}]$ is used as a physical condition in PA-DSC, enabling the correction module to distinguish feasible PV generation intervals from night-time regimes.

\subsubsection{Gated distribution shift correction}

After constructing the power sequence, TOPSIS-based weather score, timestamp feature, and day/night indicator, PA-DSC performs a gated post-prediction correction on the target-power forecast. Let $\hat{\mathbf{y}}^{\mathrm{pre}}_{t}\in\mathbb{R}^{H}$ denote the preliminary target-power prediction. For each forecasting step, we construct the input of the PA-DSC module as:
\begin{equation}
    \mathbf{C}_{t}
    =
    \left[
    \mathbf{x}^{\mathrm{pow}}_{t},
    \mathbf{q}_{t},
    \boldsymbol{\tau}_{t},
    \mathbf{z}_{t}
    \right]
    \in
    \mathbb{R}^{H\times 4},
\end{equation}
where $\mathbf{x}^{\mathrm{pow}}_{t}$ is the aligned recent power sequence, $\mathbf{q}_{t}$ is the weather score, $\boldsymbol{\tau}_{t}$ is the timestamp feature, and $\mathbf{z}_{t}$ is the day/night indicator. The matrix is first projected into a hidden space and then processed by a compact dilated temporal convolutional network:
\begin{equation}
    \mathbf{H}^{(0)}_{t}
    =
    \mathrm{Conv}_{1\times 1}
    \left(
    \mathbf{C}_{t}^{\top}
    \right),
\end{equation}
\begin{equation}
    \mathbf{H}^{(\ell)}_{t}
    =
    \mathbf{H}^{(\ell-1)}_{t}
    +
    \mathcal{F}_{\ell}
    \left(
    \mathbf{H}^{(\ell-1)}_{t}
    \right),
    \quad
    \ell=1,\ldots,L_{\mathrm{dsc}},
\end{equation}
where $\mathcal{F}_{\ell}(\cdot)$ denotes a depthwise separable temporal convolution block. This design allows PA-DSC to capture local and multi-scale correction patterns while keeping the correction module lightweight. The output head first produces three unconstrained latent factors:
\begin{equation}
    \mathbf{h}^{\mu}_t,\mathbf{h}^{s}_t,\mathbf{h}^{g}_t
    =
    \operatorname{Head}
    \left(
    \mathbf{H}^{(L_{\mathrm{dsc}})}_t
    \right).
\end{equation}

These factors are transformed as:
\begin{equation}
    \boldsymbol{\mu}_t
    =
    \tanh\left(\mathbf{h}^{\mu}_t\right),
    \qquad
    \mathbf{s}_t
    =
    \sigma\left(\mathbf{h}^{s}_t\right),
    \qquad
    \mathbf{g}_t
    =
    \sigma\left(\mathbf{h}^{g}_t\right),
\end{equation}
where $\boldsymbol{\mu}_{t}\in\mathbb{R}^{H}$ is a mean-shift term, $\mathbf{s}_{t}\in\mathbb{R}^{H}$ is a scale-adjustment factor, $\mathbf{g}_{t}\in\mathbb{R}^{H}$ is a correction gate, and $\sigma(\cdot)$ denotes the sigmoid function. PA-DSC forms a distribution-corrected candidate:
\begin{equation}
    \tilde{\mathbf{y}}^{\mathrm{dsc}}_{t}
    =
    \mathbf{s}_{t}
    \odot
    \hat{\mathbf{y}}^{\mathrm{pre}}_{t}
    +
    \boldsymbol{\mu}_{t},
\end{equation}
and then blends it with the preliminary prediction through the learned gate:
\begin{equation}
    \bar{\mathbf{y}}^{\mathrm{dsc}}_{t}
    =
    \hat{\mathbf{y}}^{\mathrm{pre}}_{t}
    +
    \mathbf{g}_{t}
    \odot
    \left(
    \tilde{\mathbf{y}}^{\mathrm{dsc}}_{t}
    -
    \hat{\mathbf{y}}^{\mathrm{pre}}_{t}
    \right).
\end{equation}

The correction is applied selectively. The gate allows PA-DSC to revise the forecast only when the learned physical and temporal context indicates that a distributional correction is needed.

\subsection{Physics-constrained learning loss function}

Common point-forecasting losses, such as MAE and MSE, assign the same weight to every prediction error. This assumption is not well suited to PV power generation, whose behavior changes markedly across regimes defined by operating conditions. As a result, the training data are often imbalanced across physical regimes. Regular generation periods typically comprise the majority of the dataset, while nighttime, peak-generation, and ramping-transition periods occur less frequently but remain crucial for reliable operation. A uniform error objective may bias the model toward the dominant regular regime and weaken its ability to handle low-power, high-power, or rapidly changing conditions. To this end, we introduce a physics-constrained learning loss that divides the target sequence by PV operating regime and adaptively reweights the forecasting errors. This encourages the model to learn both the average temporal pattern and the physical behavior.

To make the loss function consistent with PV operating conditions, the target sequence is first interpreted in the physical power scale. For a normalized target value $y_{i,h}$, its physical-scale value is obtained by $y^{\mathrm{phy}}_{i,h} = y_{i,h}\sigma_{\mathrm{data}} + \mu_{\mathrm{data}}$, where $\mu_{\mathrm{data}}$ and $\sigma_{\mathrm{data}}$ denote the dataset-level mean and standard deviation of the target power. Based on $y^{\mathrm{phy}}_{i,h}$, we partition each valid prediction point into four PV operating regimes. 

We categorize each target value into different physical regimes according to its physical-scale PV power $y^{\mathrm{phy}}_{i,h}$. Specifically, the night or low-power regime is identified as $\mathcal{M}^{\mathrm{low}}_{i,h}=\mathbb{I}(y^{\mathrm{phy}}_{i,h}\le \tau_{\mathrm{low}})$, while the peak-generation regime is defined as $\mathcal{M}^{\mathrm{peak}}_{i,h}=\mathbb{I}(y^{\mathrm{phy}}_{i,h}> \tau_{\mathrm{peak}})$. To capture rapid changes in PV output, the physical ramp magnitude is calculated as $\Delta y^{\mathrm{phy}}_{i,h}=|y^{\mathrm{phy}}_{i,h}-y^{\mathrm{phy}}_{i,h-1}|$, and the ramping regime is further defined as $\mathcal{M}^{\mathrm{ramp}}_{i,h}
= \mathbb{I}\left( \Delta y^{\mathrm{phy}}_{i,h} \ge \tau_{\mathrm{ramp}} \ \mathrm{and}\ \mathcal{M}^{\mathrm{low}}_{i,h}=0 \ \mathrm{and}\ \mathcal{M}^{\mathrm{peak}}_{i,h}=0 \right)$. Here, $\tau_{\mathrm{low}}$, $\tau_{\mathrm{peak}}$, and $\tau_{\mathrm{ramp}}$ are data-driven thresholds estimated from the training set. To ensure mutually exclusive regime assignments, peak-generation points are first identified, followed by ramping points and low-power points, while the remaining valid samples are regarded as regular-generation points.

The thresholds used for regime partition are estimated only from the training split in the physical power scale and are then kept fixed during model training and evaluation. Let
$\mathcal{Y}^{\mathrm{tr}}=\{y^{\mathrm{phy}}_n\}_{n=1}^{N_{\mathrm{tr}}}$
denote the target-power values in the training set, and let
$Q_{\alpha}(\cdot)$ denote the empirical $\alpha$-quantile. The low-power threshold is obtained from the lower tail of positive power values:
\begin{equation}
    \tau_{\mathrm{low}}
    =
    \max
    \left(
    Q_{0.01}
    \left(
    \{y \in \mathcal{Y}^{\mathrm{tr}} \mid y>0\}
    \right),
    \epsilon_0
    \right),
\end{equation}
where $\epsilon_0$ is a small constant. The peak-generation threshold is computed from non-low-power samples:
\begin{equation}
    \tau_{\mathrm{peak}}
    =
    Q_{0.90}
    \left(
    \{y \in \mathcal{Y}^{\mathrm{tr}} \mid y>\tau_{\mathrm{low}}\}
    \right).
\end{equation}

For ramping states, we first calculate the absolute temporal variation in the training target sequence,
\begin{equation}
    d_n
    =
    \left|
    y^{\mathrm{phy}}_{n}
    -
    y^{\mathrm{phy}}_{n-1}
    \right|,
\end{equation}
and define the ramp threshold as:
\begin{equation}
    \tau_{\mathrm{ramp}}
    =
    Q_{0.80}
    \left(
    \{d_n \mid d_n>0\}
    \right).
\end{equation}

These data-driven thresholds allow the loss function to adapt to the power range and variability of each PV farm, rather than relying on manually specified operating boundaries.

After the physical regimes are determined, we construct adaptive segment weights according to the sample frequency of each regime. Let
$\Omega$ denote the set of valid prediction points in a mini-batch, and let
$\Omega_k$ denotes the subset belonging to the regime
$k\in\{\mathrm{peak},\mathrm{ramp},\mathrm{low},\mathrm{regular}\}$.
The number of valid samples and regime-specific samples is denoted as $N = |\Omega|$ and $N_k = |\Omega_k|$.

For each regime, an unnormalized weight is computed as:
\begin{equation}
    \bar{w}_k
    =
    \sqrt{
    \frac{N}{N_k+\epsilon}
    },
\end{equation}
where $\epsilon$ avoids numerical instability when a regime contains very few samples. Each prediction point inherits the weight of its assigned regime:
\begin{equation}
    \bar{w}_{i,h}
    =
    \sum_{k}
    \bar{w}_k
    \mathbb{I}
    \left(
    (i,h)\in\Omega_k
    \right).
\end{equation}

Finally, the weights are normalized by their mini-batch mean:
\begin{equation}
    w_{i,h}
    =
    \frac{
    \bar{w}_{i,h}
    }{
    \frac{1}{N}
    \sum_{(i,h)\in\Omega}
    \bar{w}_{i,h}
    }.
\end{equation}

Given the adaptive regime weights, the physics-constrained point forecasting loss is defined as a weighted MAE over all valid prediction points:
\begin{equation}
    \mathcal{L}_{\mathrm{phy}}
    =
    \frac{1}{|\Omega|}
    \sum_{(i,h)\in\Omega}
    w_{i,h}
    \left|
    \hat{y}_{i,h}
    -
    y_{i,h}
    \right|,
\end{equation}
where $\hat{y}{i,h}$ and $y{i,h}$ denote the predicted and ground-truth target power values, respectively, and $w_{i,h}$ is the adaptive weight determined by the physical regime of $y_{i,h}$. The weights are estimated from regime statistics in the physical scale and then applied to the training error. In this way, the loss keeps the standard regression objective while placing greater optimization weight on physically important PV states.

For probabilistic forecasting, PARA-PV predicts a set of quantile outputs
$\{\hat{y}^{(q)}_{i,h}\}_{q\in\mathcal{Q}}$, where $\mathcal{Q}$ denotes the predefined quantile levels. The probabilistic objective is formulated as the average quantile loss:
\begin{equation}
    \mathcal{L}_{\mathrm{prob}}
    =
    \frac{1}{|\Omega||\mathcal{Q}|}
    \sum_{(i,h)\in\Omega}
    \sum_{q\in\mathcal{Q}}
    \rho_q
    \left(
    y_{i,h}
    -
    \hat{y}^{(q)}_{i,h}
    \right),
\end{equation}
where $\rho_q(\cdot)$ is the pinball loss function $\rho_q(e) = \max \left( q e, (q-1)e \right)$, $e=y_{i,h}-\hat{y}^{(q)}_{i,h}$ denotes the quantile prediction error. This objective aims to encourage the predicted quantiles to approximate the conditional predictive distribution of PV power, thus supporting uncertainty-aware PV forecasting.

\section{Experimental verification}

\subsection{Experiment setting}

\textbf{Data selection}: In this study, we selected two different PV power stations with distinct geographical locations and installed capacities. The data were obtained from the publicly available PV dataset released for the Chinese State Grid Renewable Energy Generation Forecasting Competition \cite{chen2022solar}. Specifically, Dataset 1 has an installed capacity of 50 MW and contains records for the full year of 2019, whereas Dataset 2 has an installed capacity of 35 MW and covers the two years from 2019 to 2020. Dataset 1 contains 35,040 data points, and Dataset 2 contains 70,176 data points, both recorded at a 15-minute resolution. The datasets include historical power generation and relevant meteorological features. Missing values were imputed using a moving-window method. The training, validation, and test sets were split at a ratio of 8:1:1.

\textbf{Evaluation metrics}: To comprehensively assess the proposed model, we employ metrics for both point and probabilistic forecasting. All metrics reported in subsequent experiments are calculated on the standardized target scale. The standardization parameters are estimated exclusively from the corresponding training set and applied unchanged to the validation and test sets. For point forecasting, we adopt four widely used metrics: mean squared error (MSE), mean absolute error (MAE), root mean squared error (RMSE), and the coefficient of determination ($R^2$). These metrics are defined as follows \cite{tian2026enhancing}:
\begin{align}
\mathrm{MSE} & = \frac{1}{N}\sum_{i=1}^{N}(y_i-\hat{y}_i)^2 \notag \\ 
\mathrm{MAE} & = \frac{1}{N}\sum_{i=1}^{N}|y_i-\hat{y}_i| \notag \\
\mathrm{RMSE} & = \sqrt{\frac{1}{N}\sum_{i=1}^{N}(y_i-\hat{y}_i)^2} \notag \\
R^2 & =1-\frac{\sum_{i=1}^{N}(y_i-\hat{y}_i)^2}{\sum_{i=1}^{N}(y_i-\bar{y})^2}  \notag
\end{align}
where $y_i$ and $\hat{y}_i$ denote the observed and predicted PV power values, $\bar{y}$ is the mean of the observations, and $N$ is the number of samples. For probabilistic forecasting, the prediction interval coverage probability, prediction interval normalized average width, and average quantile loss are used to assess the reliability, sharpness, and quantile accuracy of the interval. They are formulated as: 
\begin{align}
\mathrm{PICP} & =\frac{1}{N}\sum_{i=1}^{N}\mathbb{I}(L_i\le y_i\le U_i) \notag \\
\mathrm{PINAW} & = \frac{1}{NR}\sum_{i=1}^{N}(U_i-L_i) \notag \\
\mathrm{AQL}  & = \frac{1}{N|\mathcal{Q}|} \sum_{i=1}^{N} \sum_{q\in\mathcal{Q}} \rho_q(y_i-\hat{y}_{i}^{q}) \notag
\end{align}
where $L_i$ and $U_i$ are the lower and upper bounds of the prediction interval, $R$ denotes the range of the observed values, $\mathcal{Q}$ is the set of quantile levels, $\hat{y}_{i}^{q}$ is the predicted $q$-th quantile, and $\rho_q(e)=e(q-\mathbb{I}(e<0))$ is the quantile loss function. Lower values of MSE, MAE, RMSE, PINAW, and AQL indicate better performance, while a higher $R^2$ and a PICP closer to the nominal confidence level are preferred.

\vspace{0.1\baselineskip}

\textbf{Baseline models}: To comprehensively evaluate the forecasting performance of PARA-PV, we compare it with seven representative baseline models covering recurrent networks, Transformer-based architectures, linear models, period-aware models, and LLM-based forecasting paradigms. Specifically, \textbf{LSTM} is a classical recurrent neural network that captures temporal dependencies through gated memory mechanisms \cite{hochreiter1997long}. \textbf{Informer} is an efficient Transformer-based model designed for long-sequence forecasting by reducing the computational cost of self-attention \cite{zhou2021informer}. \textbf{DLinear} decomposes the input series into trend and seasonal components and performs forecasting with simple linear mappings \cite{zeng2023transformers}. \textbf{iTransformer} treats variables as tokens and applies attention across variables to model multivariate dependencies \cite{liu2024itransformer}. \textbf{TimesNet} transforms one-dimensional time series into two-dimensional period-based representations to capture multi-period temporal patterns \cite{wu2023timesnet}. \textbf{TimeLLM} adapts pre-trained large language models to time-series forecasting by reprogramming numerical sequences into LLM-compatible representations \cite{jin2024time}. \textbf{TimeVLM} introduces vision-language modeling into time-series forecasting by converting temporal patterns into multimodal representations and leveraging pre-trained vision-language knowledge \cite{zhong2025time}.

\vspace{0.1\baselineskip}

\textbf{Parameter design}: Hyperparameter settings are critical for ensuring fair comparisons, reproducibility, and stable model training. In this study, the main experimental configurations are kept consistent across the proposed PARA-PV and all baseline models whenever applicable, including the input length, forecasting horizons, batch size, and number of training epochs. Model-specific hyperparameters are then configured according to the architectural characteristics and requirements of each method. The detailed parameter settings of PARA-PV and all baseline models are provided in \textbf{\ref{appendix.A}}.

\vspace{0.1\baselineskip}

\textbf{Operation environment}: All experiments were conducted on a workstation equipped with an Intel i7-9700 CPU running at 3.00 GHz, 16 GB of RAM, and two NVIDIA GeForce RTX 3090 GPUs. The proposed model was implemented in PyTorch, with PyCharm used as the development environment.

\subsection{Point forecasting comparative model study}

\begin{figure}[!h]
    \centering
    \includegraphics[width=0.9\linewidth]{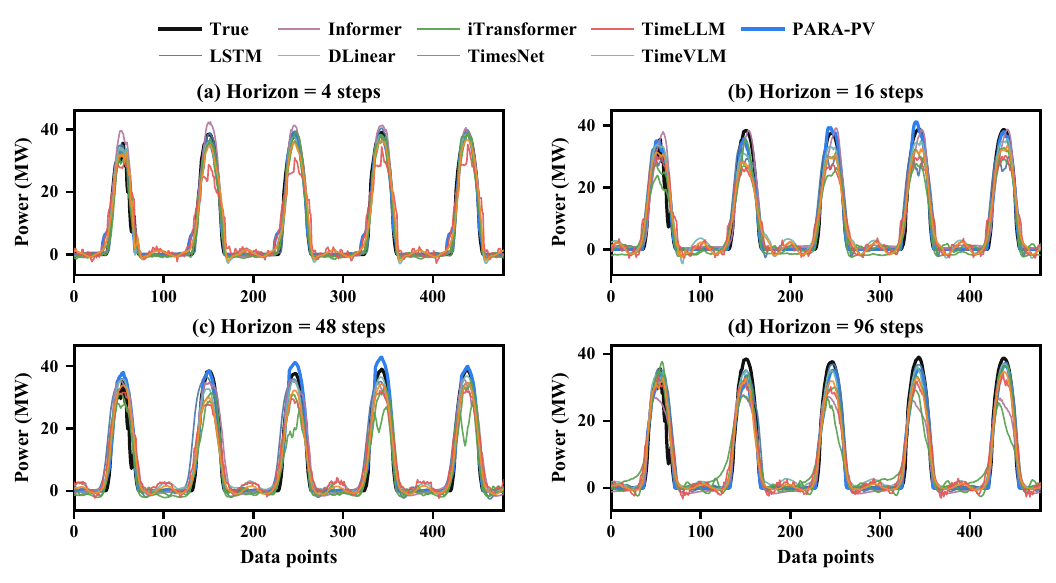}
    \vspace{-0.6\baselineskip}
    \caption{The results of point forecasting in Data 1 (50 MW).}
    \label{fig:Data1-point}
\end{figure}

\begin{figure}[!h]
    \centering
    \includegraphics[width=0.9\linewidth]{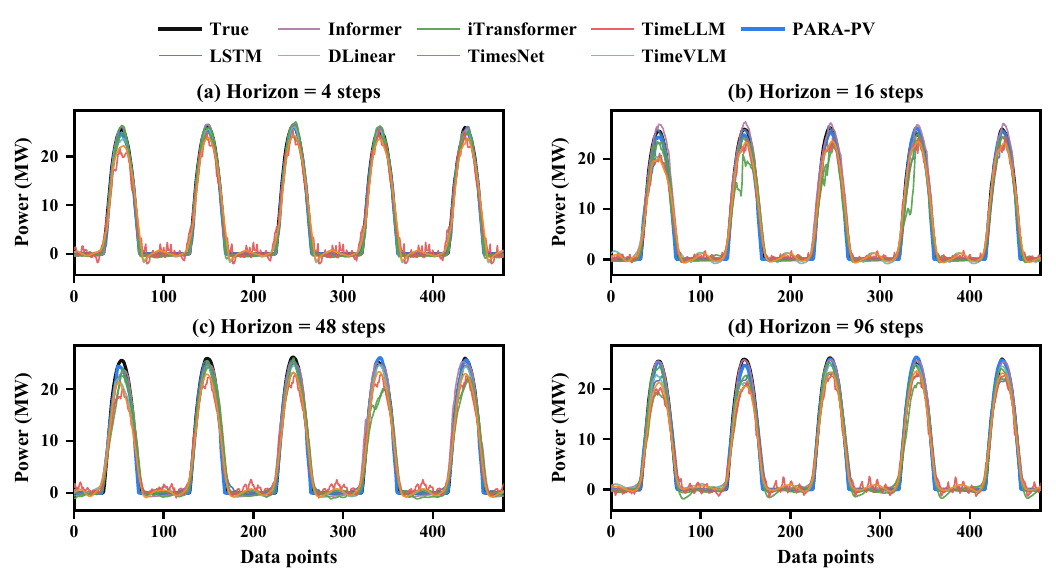}
    \vspace{-0.6\baselineskip}
    \caption{The results of point forecasting in Data 2 (35 MW).}
    \label{fig:Data2-point}
\end{figure}

To rigorously evaluate the performance of the proposed model, we conduct a comprehensive comparative study on two distinct PV power datasets, namely Data 1 (50 MW) and Data 2 (35 MW). The evaluation spans multiple prediction horizons, including 4, 16, 48, and 96 steps ahead to verify the robustness of models across short-term and long-term forecasting scenarios. We benchmark PARA-PV against seven state-of-the-art baselines encompassing classical recurrent networks, advanced transformers, and large language model adaptations. Performance is quantified using four standard evaluation metrics, including MSE, MAE, RMSE, and the $R^2$. The results of the comparison are shown in \textbf{Fig.~\ref{fig:Data1-point}}, and \textbf{Fig.~\ref{fig:Data2-point}}.

The experimental results for Data 1, representing a 50 MW capacity system, are summarized in \textbf{Table~\ref{tab:point_model_comparison}}. A critical observation is that while LSTM demonstrates the top performance at the shortest forecasting horizon of 4 steps with an MSE of 0.0384 and an $R^2$ of 0.9493, its accuracy degrades sharply as the prediction window elongates, with the MSE surging to 0.2887 at horizon 48. In contrast, PARA-PV maintains remarkable stability and establishes clear superiority in all extended horizons. At the longest horizon of 96 steps, PARA-PV achieves an MSE of 0.1272 and an $R^2$ of 0.8305, outperforming the second-best baseline, TimeVLM, which obtains an MSE of 0.1433 and an $R^2$ of 0.8091. Looking at the overall average performance, PARA-PV delivers the lowest average MSE of 0.0962 and the highest $R^2$ of 0.8723, marking a substantial advancement over modern transformer-based architectures and LLM-tailored variants like TimeLLM, which struggles significantly with an average MSE of 0.1652.

\begin{table}[!h]
\centering
\setlength{\tabcolsep}{4pt}
\renewcommand{\arraystretch}{1.0}
\captionsetup{width=\textwidth}
\caption{Point forecasting comparison across multiple prediction horizons in Data 1.}
\label{tab:point_model_comparison}

\resizebox{\textwidth}{!}{
\begin{tabular}{llcccccccc}
\toprule \hline
\multirow{2}{*}{Horizon} 
& \multirow{2}{*}{Metric}
& \multicolumn{8}{c}{Model} \\
\cmidrule(lr){3-10}
& 
& LSTM & Informer & DLinear & iTransformer & TimesNet & TimeLLM & TimeVLM & \textbf{PARA-PV} \\
\midrule

\multirow{4}{*}{4}
& MSE   & \textbf{0.0384}& 0.0459& 0.0491& 0.0488& 0.0505& 0.1455& 0.0519& \underline{0.0458} \\
& MAE   & \textbf{0.0802}& 0.1050& 0.1180& 0.1232& 0.1258& 0.2435& 0.1239& \underline{0.0962} \\
& RMSE  & \textbf{0.1960}& 0.2144& 0.2215& 0.2208& 0.2247& 0.3815& 0.2279& \underline{0.2140} \\
& R$^2$ & \textbf{0.9493}& 0.9394& 0.9352& 0.9357& 0.9334& 0.8079& 0.9315& \underline{0.9396} \\
\midrule

\multirow{4}{*}{16}
& MSE   & 0.1364& 0.1586& 0.1224& 0.1179& \underline{0.1148}& 0.1783& 0.1198& \textbf{0.0949} \\
& MAE   & \underline{0.1921}& 0.1946& 0.2208& 0.2157& 0.2157& 0.2888& 0.2075& \textbf{0.1343} \\
& RMSE  & 0.3693& 0.3982& 0.3499& 0.3433& \underline{0.3388}& 0.4222& 0.3462& \textbf{0.3081} \\
& R$^2$ & 0.8193& 0.7899& 0.8378& 0.8438& \underline{0.8479}& 0.7638& 0.8412& \textbf{0.8742} \\
\midrule

\multirow{4}{*}{48}
& MSE   & 0.2887& 0.1887& 0.1540& 0.1425& 0.1615& 0.1769& \underline{0.1413}& \textbf{0.1168} \\
& MAE   & 0.3023& 0.2258& 0.2660& 0.2230& 0.2304& 0.2726& \underline{0.2207}& \textbf{0.1526} \\
& RMSE  & 0.5373& 0.4344& 0.3924& 0.3775& 0.4018& 0.4206& \underline{0.3759}& \textbf{0.3418} \\
& R$^2$ & 0.6170& 0.7496& 0.7958& 0.8109& 0.7858& 0.7654& \underline{0.8125}& \textbf{0.8450} \\
\midrule

\multirow{4}{*}{96}
& MSE   & 0.1560& 0.1693& 0.1586& 0.1595& 0.1731& 0.1602& \underline{0.1433}& \textbf{0.1272} \\
& MAE   & 0.2246& 0.2417& 0.2746& 0.2356& 0.2473& 0.2530& \underline{0.2230}& \textbf{0.1575} \\
& RMSE  & 0.3950& 0.4114& 0.3982& 0.3993& 0.4161& 0.4003& \underline{0.3786}& \textbf{0.3567} \\
& R$^2$ & 0.7921& 0.7744& 0.7887& 0.7875& 0.7693& 0.7865& \underline{0.8091}& \textbf{0.8305} \\
\midrule

\rowcolor{mediumgray}
& MSE $\downarrow$ & 0.1549& 0.1406& 0.1210& 0.1172& 0.1250& 0.1652& \underline{0.1141}& \textbf{0.0962} \\
\rowcolor{mediumgray}
& MAE $\downarrow$ & 0.1998& \underline{0.1918}& 0.2198& 0.1994& 0.2048& 0.2645& 0.1938& \textbf{0.1352} \\
\rowcolor{mediumgray}
& RMSE $\downarrow$  & 0.3744& 0.3646& 0.3405& 0.3352& 0.3454& 0.4062& \underline{0.3322}& \textbf{0.3052} \\
\rowcolor{mediumgray}
\multirow{-4}{*}{\textbf{AVG}} & R$^2$ $\uparrow$ & 0.7944& 0.8133& 0.8394& 0.8445& 0.8341& 0.7809& \underline{0.8486}& \textbf{0.8723} \\
\hline

\bottomrule
\end{tabular}
}
\vspace{4pt}
\begin{minipage}{\textwidth}
\raggedright
\footnotesize{
Note: The best results are shown in \textbf{BOLD}, and the second-best results are \underline{UNDERLINED}.
}
\end{minipage}
\end{table}

\textbf{Table~\ref{tab:point_model_comparison_2}} presents the comparative evaluation on Data 2, a 35 MW capacity system, which further substantiates the architectural benefits of the proposed framework. Similar to the trends observed in the first dataset, LSTM excels primarily at the ultra-short horizon of 4 steps, capturing an MSE of 0.0268, whereas PARA-PV secures the best MAE of 0.0637. However, as the horizon scales to 16, 48, and 96 steps, PARA-PV consistently dominates all metrics. For instance, at horizon 96, PARA-PV compresses the MSE to 0.1000 and elevates the $R^2$ to 0.8958, outperforming the second-best baseline, DLinear, which reports an MSE of 0.1143 and an $R^2$ of 0.8809. Interestingly, LSTM remains the second-best model on average for Data 2, achieving a mean MSE of 0.0788 and an $R^2$ of 0.9179, indicating that simpler sequential or linear structures can occasionally outperform overly complex representations like TimeLLM in specific capacity contexts, yet they still fall short of the robust generalized representations captured by PARA-PV, which secures an average MSE of 0.0689 and an $R^2$ of 0.9282.

\begin{table}[!h]
\centering
\setlength{\tabcolsep}{4pt}
\renewcommand{\arraystretch}{1.0}
\captionsetup{width=\textwidth}
\caption{Point forecasting comparison across multiple prediction horizons in Data 2.}
\label{tab:point_model_comparison_2}

\resizebox{\textwidth}{!}{
\begin{tabular}{llcccccccc}
\toprule \hline
\multirow{2}{*}{Horizon} 
& \multirow{2}{*}{Metric}
& \multicolumn{8}{c}{Model} \\
\cmidrule(lr){3-10}
& 
& LSTM & Informer & DLinear & iTransformer & TimesNet & TimeLLM & TimeVLM & \textbf{PARA-PV} \\
\midrule

\multirow{4}{*}{4}
& MSE   & \textbf{0.0268} & 0.0286 & 0.0320 & 0.0292 & 0.0316 & 0.1025 & 0.0382 & \underline{0.0275} \\
& MAE   & \underline{0.0651} & 0.0882 & 0.0933 & 0.0859 & 0.0951 & 0.2128 & 0.1139 & \textbf{0.0637} \\
& RMSE  & \textbf{0.1637} & 0.1690 & 0.1788 & 0.1708 & 0.1778 & 0.3202 & 0.1954 & \underline{0.1659} \\
& R$^2$ & \textbf{0.9720} & 0.9701 & 0.9666 & 0.9695 & 0.9669 & 0.8928 & 0.9601 & \underline{0.9712} \\
\midrule

\multirow{4}{*}{16}
& MSE   & \underline{0.0678} & 0.1135 & 0.0818 & 0.0787 & 0.0744 & 0.1013 & 0.0807 & \textbf{0.0601} \\
& MAE   & \underline{0.1157} & 0.1736 & 0.1757 & 0.1485 & 0.1398 & 0.1878 & 0.1630 & \textbf{0.0977} \\
& RMSE  & \underline{0.2604} & 0.3369 & 0.2861 & 0.2806 & 0.2728 & 0.3182 & 0.2841 & \textbf{0.2451} \\
& R$^2$ & \underline{0.9292} & 0.8814 & 0.9145 & 0.9178 & 0.9223 & 0.8942 & 0.9157 & \textbf{0.9372} \\
\midrule

\multirow{4}{*}{48}
& MSE   & \underline{0.1007} & 0.1299 & 0.1070 & 0.1085 & 0.1095 & 0.1285 & 0.1138 & \textbf{0.0878} \\
& MAE   & \underline{0.1671} & 0.1878 & 0.1920 & 0.1802 & 0.1776 & 0.2175 & 0.1974 & \textbf{0.1212} \\
& RMSE  & \underline{0.3173} & 0.3604 & 0.3271 & 0.3294 & 0.3310 & 0.3585 & 0.3374 & \textbf{0.2963} \\
& R$^2$ & \underline{0.8950} & 0.8645 & 0.8884 & 0.8868 & 0.8858 & 0.8660 & 0.8813 & \textbf{0.9084} \\
\midrule

\multirow{4}{*}{96}
& MSE   & 0.1198 & 0.1151 & \underline{0.1143} & 0.1169 & 0.1169 & 0.1306 & 0.1235 & \textbf{0.1000} \\
& MAE   & 0.1895 & \underline{0.1730} & 0.1984 & 0.2004 & 0.1980 & 0.2221 & 0.2076 & \textbf{0.1228} \\
& RMSE  & 0.3461 & 0.3393 & \underline{0.3381} & 0.3420 & 0.3420 & 0.3614 & 0.3515 & \textbf{0.3163} \\
& R$^2$ & 0.8753 & 0.8801 & \underline{0.8809} & 0.8782 & 0.8782 & 0.8640 & 0.8713 & \textbf{0.8958} \\
\midrule

\rowcolor{mediumgray}
& MSE $\downarrow$ & \underline{0.0788} & 0.0968 & 0.0838 & 0.0833 & 0.0831 & 0.1157 & 0.0890 & \textbf{0.0689} \\
\rowcolor{mediumgray}
& MAE $\downarrow$ & \underline{0.1343} & 0.1557 & 0.1648 & 0.1537 & 0.1526 & 0.2100 & 0.1705 & \textbf{0.1014} \\
\rowcolor{mediumgray}
& RMSE $\downarrow$  & \underline{0.2719} & 0.3014 & 0.2825 & 0.2807 & 0.2809 & 0.3396 & 0.2921 & \textbf{0.2559} \\
\rowcolor{mediumgray}
\multirow{-4}{*}{\textbf{AVG}} & R$^2$ $\uparrow$ & \underline{0.9179} & 0.8990 & 0.9126 & 0.9131 & 0.9133 & 0.8792 & 0.9071 & \textbf{0.9282} \\
\hline

\bottomrule
\end{tabular}
}

\vspace{4pt}
\begin{minipage}{\textwidth}
\raggedright
\footnotesize{
Note: The best results are shown in \textbf{BOLD}, and the second-best results are \underline{UNDERLINED}.
}
\end{minipage}
\end{table}

In summary, the experimental results yield the following conclusions. First, local temporal dependencies captured by recurrent structures are highly effective for immediate, short-term predictions, but they fail to sustain accuracy over broader horizons due to error accumulation. Second, while sophisticated large-scale models like TimeLLM and TimeVLM introduce powerful cross-domain representations, their performance heavily depends on the underlying data scale and capacity characteristics, which can occasionally lead to suboptimal results in specialized regression tasks. Ultimately, PARA-PV successfully bridges this gap by striking a balance between immediate precision and long-term stability. It consistently yields the lowest average errors and highest variance explanation across diverse capacities, establishing itself as a highly reliable and robust solution for multi-horizon PV power forecasting.

\subsection{Probabilistic forecasting comparative model study}

To comprehensively assess the capability of the models in uncertainty quantification, this section presents a comparative study on probabilistic forecasting. The evaluation is conducted across multiple prediction horizons, including 4, 16, 48, and 96 steps ahead on both Data 1 and Data 2. We analyze the performance of the proposed PARA-PV model against seven prominent baselines: LSTM, Informer, DLinear, iTransformer, TimesNet, TimeLLM, and TimeVLM. The models are evaluated using three standard probabilistic metrics: Prediction Interval Coverage Probability (PICP) and Prediction Interval Normalized Average Width (PINAW) at both 80\% and 90\% confidence levels, alongside Average Quantile Loss (AQL), providing a holistic overview of interval reliability and sharpness.

The experimental results for Data 1 are detailed in \textbf{Table~\ref{tab:prob_model_comparison}}, illuminating distinct behaviors across different horizons. At the shortest horizon of 4 steps, LSTM shows strong initial probabilistic calibration with a PICP$_{80}$ of 0.9071 and an AQL of 0.0239, while TimeVLM achieves the top PICP$_{90}$ of 0.9563. However, as the prediction horizon extends to 48 and 96 steps, PARA-PV demonstrates an exceptional capability to optimize interval sharpness without severe under-coverage. Specifically, at horizon 96, PARA-PV achieves the narrowest interval widths with a PINAW$_{80}$ of 0.1583 and a PINAW$_{90}$ of 0.1813, substantially outperforming baselines like TimeVLM, which exhibits an over-conservative PINAW$_{90}$ of 0.3810 despite its high PICP$_{90}$ of 0.9648. On average, DLinear secures the lowest overall AQL of 0.0374, whereas PARA-PV achieves a competitive average AQL of 0.0417, matching iTransformer while maintaining the lowest average interval widths of 0.1303 for PINAW$_{80}$ and 0.1625 for PINAW$_{90}$.

\begin{table}[!h]
\centering
\setlength{\tabcolsep}{4pt}
\renewcommand{\arraystretch}{1.0}
\captionsetup{width=\textwidth}
\caption{Probabilistic forecasting comparison across multiple prediction horizons in Data 1.}
\label{tab:prob_model_comparison}
\resizebox{\linewidth}{!}{
\begin{tabular}{llcccccccc}
\toprule \hline
\multirow{2}{*}{Horizon}
& \multirow{2}{*}{Metric}
& \multicolumn{8}{c}{Model} \\
\cmidrule(lr){3-10}
&
& LSTM & Informer & DLinear & iTransformer & TimesNet & TimeLLM & TimeVLM & \textbf{PARA-PV} \\
\midrule

\multirow{5}{*}{4}
& PICP$_{80}$ & \textbf{0.9071} & 0.4374 & 0.8530 & 0.7140 & 0.6340 & 0.8590 & \underline{0.9003} & 0.8258 \\
& PINAW$_{80}$ & 0.0878 & 0.0853 & 0.0901 & \textbf{0.0819} & \underline{0.0851} & 0.2317 & 0.1412 & 0.0881 \\
& PICP$_{90}$ & \underline{0.9519} & 0.6420 & 0.9104 & 0.9116 & 0.8536 & 0.9474 & \textbf{0.9563} & 0.8862 \\
& PINAW$_{90}$ & 0.1257 & \textbf{0.1213} & 0.1251 & \underline{0.1217} & 0.1218 & 0.4089 & 0.2311 & 0.1219 \\
& AQL & \textbf{0.0239} & 0.0278 & \underline{0.0259} & 0.0272 & 0.0284 & 0.0547 & 0.0347 & 0.0290 \\
\midrule

\multirow{5}{*}{16}
& PICP$_{80}$ & 0.7793 & 0.3885 & 0.7796 & 0.5932 & 0.4601 & \underline{0.7957} & 0.7793 & \textbf{0.8497} \\
& PINAW$_{80}$ & 0.1750 & 0.1522 & 0.1347 & \underline{0.1294} & \textbf{0.1277} & 0.2724 & 0.1627 & 0.1306 \\
& PICP$_{90}$ & 0.9208 & 0.5789 & 0.8919 & 0.7957 & 0.7354 & \textbf{0.9382} & \underline{0.9232} & 0.8906 \\
& PINAW$_{90}$ & 0.2434 & 0.1994 & 0.1916 & \underline{0.1822} & 0.1974 & 0.4347 & 0.2502 & \textbf{0.1604} \\
& AQL & 0.0491 & 0.0428 & \textbf{0.0383} & 0.0455 & 0.0486 & 0.0645 & 0.0464 & \underline{0.0426} \\
\midrule

\multirow{5}{*}{48}
& PICP$_{80}$ & 0.7751 & 0.4540 & 0.8223 & 0.6923 & 0.7434 & \underline{0.8231} & \textbf{0.8657} & 0.8166 \\
& PINAW$_{80}$ & 0.2360 & 0.1682 & 0.1984 & \underline{0.1600} & 0.1814 & 0.3192 & 0.2613 & \textbf{0.1442} \\
& PICP$_{90}$ & 0.8982 & 0.7160 & \underline{0.9566} & 0.8785 & 0.9014 & 0.9305 & \textbf{0.9689} & 0.8866 \\
& PINAW$_{90}$ & 0.3224 & 0.2240 & 0.2746 & \underline{0.2226} & 0.2453 & 0.4306 & 0.3621 & \textbf{0.1863} \\
& AQL & 0.0664 & 0.0495 & \textbf{0.0415} & 0.0475 & 0.0491 & 0.0623 & 0.0498 & \underline{0.0460} \\
\midrule

\multirow{5}{*}{96}
& PICP$_{80}$ & 0.7825 & 0.7670 & 0.8219 & 0.7613 & 0.7593 & 0.8157 & \underline{0.8366} & \textbf{0.8518} \\
& PINAW$_{80}$ & 0.2486 & \underline{0.2178} & 0.2265 & 0.2175 & 0.2566 & 0.3488 & 0.2856 & \textbf{0.1583} \\
& PICP$_{90}$ & 0.9330 & 0.9155 & \underline{0.9607} & 0.9353 & 0.9383 & 0.9210 & \textbf{0.9648} & 0.8852 \\
& PINAW$_{90}$ & 0.3183 & \underline{0.2762} & 0.3101 & 0.2972 & 0.3413 & 0.4473 & 0.3810 & \textbf{0.1813} \\
& AQL & 0.0545 & 0.0512 & \textbf{0.0438} & \underline{0.0467} & 0.0516 & 0.0647 & 0.0523 & 0.0492 \\
\midrule

\rowcolor{mediumgray}
& PICP$_{80}$ $\uparrow$ & 0.8110 & 0.5117 & 0.8192 & 0.6902 & 0.6492 & 0.8234 & \textbf{0.8455} & \underline{0.8360} \\
\rowcolor{mediumgray}
& PINAW$_{80}$ $\downarrow$ & 0.1868 & 0.1559 & 0.1624 & \underline{0.1472} & 0.1627 & 0.2930 & 0.2127 & \textbf{0.1303} \\
\rowcolor{mediumgray}
& PICP$_{90}$ $\uparrow$ & 0.9260 & 0.7131 & 0.9299 & 0.8803 & 0.8572 & \underline{0.9343} & \textbf{0.9533} & 0.8872 \\
\rowcolor{mediumgray}
& PINAW$_{90}$ $\downarrow$ & 0.2525 & \underline{0.2052} & 0.2253 & 0.2059 & 0.2264 & 0.4304 & 0.3061 & \textbf{0.1625} \\
\rowcolor{mediumgray}
\multirow{-5}{*}{\textbf{AVG}} & AQL $\downarrow$ & 0.0485 & 0.0428 & \textbf{0.0374} & 0.0417 & 0.0444 & 0.0615 & 0.0458 & \underline{0.0417} \\
\hline

\bottomrule
\end{tabular}
}

\vspace{4pt}
\begin{minipage}{\linewidth}
\raggedright
\footnotesize{
Note: The best results are shown in \textbf{BOLD}, and the second-best results are \underline{UNDERLINED}.
}
\end{minipage}
\end{table}

\textbf{Table~\ref{tab:prob_model_comparison_2}} provides the probabilistic forecasting comparison on Data 2, further validating the efficiency of the proposed framework under a different capacity distribution. In the short-term 4-step horizon, LSTM leads with an AQL of 0.0216, followed tightly by PARA-PV at 0.0219. As the forecasting horizon scales up to long-term intervals, PARA-PV progressively outpaces the baselines in both sharpness and quantile loss optimization. At horizon 48, PARA-PV achieves the lowest AQL of 0.0365 and reduces PINAW$_{90}$ to 0.1895. Looking at the overall average performance, PARA-PV delivers the lowest average PINAW$_{80}$ of 0.1251, the lowest average PINAW$_{90}$ of 0.1718, and the minimum average AQL of 0.0326. While TimeLLM attains the highest average PICP$_{90}$ of 0.9708, its average interval width of 0.3805 implies a loss of practical resolution, highlighting the superior calibration-sharpness balance achieved by PARA-PV.

\begin{table}[!h]
\centering
\setlength{\tabcolsep}{4pt}
\renewcommand{\arraystretch}{1.0}
\captionsetup{width=\textwidth}
\caption{Probabilistic forecasting comparison across multiple prediction horizons in Data 2.}
\label{tab:prob_model_comparison_2}
\resizebox{\linewidth}{!}{
\begin{tabular}{llcccccccc}
\toprule \hline
\multirow{2}{*}{Horizon}
& \multirow{2}{*}{Metric}
& \multicolumn{8}{c}{Model} \\
\cmidrule(lr){3-10}
&
& LSTM & Informer & DLinear & iTransformer & TimesNet & TimeLLM & TimeVLM & \textbf{PARA-PV} \\
\midrule

\multirow{5}{*}{4}
& PICP$_{80}$ & \textbf{0.9321} & 0.6235 & 0.9006 & 0.6171 & \underline{0.9083} & 0.8201 & 0.9037 & 0.8857 \\
& PINAW$_{80}$ & 0.0971 & 0.0977 & 0.1092 & \textbf{0.0929} & 0.0957 & 0.2170 & 0.1400 & \underline{0.0931} \\
& PICP$_{90}$ & \textbf{0.9683} & 0.9515 & 0.9406 & 0.9452 & 0.9527 & \underline{0.9670} & 0.9665 & 0.9424 \\
& PINAW$_{90}$ & 0.1491 & 0.1432 & 0.1536 & 0.1459 & \underline{0.1382} & 0.3768 & 0.2033 & \textbf{0.1317} \\
& AQL & \textbf{0.0216} & 0.0233 & 0.0235 & 0.0228 & 0.0231 & 0.0467 & 0.0312 & \underline{0.0219} \\
\midrule

\multirow{5}{*}{16}
& PICP$_{80}$ & \textbf{0.9065} & 0.4359 & 0.8215 & 0.8305 & 0.7780 & 0.7977 & 0.7288 & \underline{0.8352} \\
& PINAW$_{80}$ & \underline{0.1265} & 0.1431 & 0.1504 & 0.1298 & 0.1280 & 0.2600 & 0.1693 & \textbf{0.1150} \\
& PICP$_{90}$ & \underline{0.9618} & 0.5310 & 0.9435 & 0.9529 & 0.8995 & \textbf{0.9686} & 0.9396 & 0.9279 \\
& PINAW$_{90}$ & \underline{0.1748} & 0.2004 & 0.2100 & 0.1838 & 0.1750 & 0.3821 & 0.2436 & \textbf{0.1616} \\
& AQL & \textbf{0.0301} & 0.0371 & 0.0323 & \underline{0.0309} & 0.0349 & 0.0512 & 0.0384 & 0.0312 \\
\midrule

\multirow{5}{*}{48}
& PICP$_{80}$ & \underline{0.8464} & 0.5587 & 0.8455 & 0.7676 & 0.7659 & \textbf{0.8534} & 0.8442 & 0.7721 \\
& PINAW$_{80}$ & 0.1691 & 0.2038 & 0.2073 & 0.1913 & \underline{0.1636} & 0.2800 & 0.2340 & \textbf{0.1241} \\
& PICP$_{90}$ & 0.9451 & 0.7696 & 0.9680 & 0.9512 & 0.9218 & \textbf{0.9753} & \underline{0.9707} & 0.9028 \\
& PINAW$_{90}$ & \underline{0.2214} & 0.2686 & 0.2766 & 0.2624 & 0.2247 & 0.3704 & 0.3067 & \textbf{0.1895} \\
& AQL & 0.0389 & 0.0451 & \underline{0.0378} & 0.0400 & 0.0418 & 0.0467 & 0.0419 & \textbf{0.0365} \\
\midrule

\multirow{5}{*}{96}
& PICP$_{80}$ & 0.7619 & 0.6550 & \underline{0.8746} & 0.7993 & 0.8119 & \textbf{0.8906} & 0.7976 & 0.7666 \\
& PINAW$_{80}$ & \underline{0.1870} & 0.2151 & 0.2217 & 0.2142 & 0.2365 & 0.3033 & 0.2443 & \textbf{0.1684} \\
& PICP$_{90}$ & 0.9261 & 0.8849 & \textbf{0.9785} & 0.9630 & 0.9659 & \underline{0.9723} & 0.9579 & 0.9173 \\
& PINAW$_{90}$ & \underline{0.2482} & 0.2857 & 0.2812 & 0.2843 & 0.3081 & 0.3929 & 0.3074 & \textbf{0.2043} \\
& AQL & 0.0440 & 0.0487 & \textbf{0.0378} & \underline{0.0405} & 0.0409 & 0.0497 & 0.0416 & 0.0409 \\
\midrule

\rowcolor{mediumgray}
& PICP$_{80}$ $\uparrow$ & \textbf{0.8617} & 0.5683 & \underline{0.8606} & 0.7536 & 0.8160 & 0.8405 & 0.8186 & 0.8149 \\
\rowcolor{mediumgray}
& PINAW$_{80}$ $\downarrow$ & \underline{0.1449} & 0.1649 & 0.1721 & 0.1571 & 0.1559 & 0.2651 & 0.1969 & \textbf{0.1251} \\
\rowcolor{mediumgray}
& PICP$_{90}$ $\uparrow$ & 0.9503 & 0.7843 & 0.9577 & 0.9531 & 0.9350 & \textbf{0.9708} & \underline{0.9587} & 0.9226 \\
\rowcolor{mediumgray}
& PINAW$_{90}$ $\downarrow$ & \underline{0.1984} & 0.2245 & 0.2303 & 0.2191 & 0.2115 & 0.3805 & 0.2652 & \textbf{0.1718} \\
\rowcolor{mediumgray}
\multirow{-5}{*}{\textbf{AVG}} & AQL $\downarrow$ & 0.0336 & 0.0386 & \underline{0.0329} & 0.0336 & 0.0352 & 0.0486 & 0.0383 & \textbf{0.0326} \\
\hline

\bottomrule
\end{tabular}
}

\vspace{4pt}
\begin{minipage}{\linewidth}
\raggedright
\footnotesize{
Note: The best results are shown in \textbf{BOLD}, and the second-best results are \underline{UNDERLINED}.
}
\end{minipage}
\end{table}

Synthesizing the empirical evidence from both tables highlights a crucial trade-off between interval coverage and sharpness in probabilistic PV power forecasting. Large language model adaptations, such as TimeLLM and TimeVLM, frequently yield higher coverage probabilities but do so by generating excessively wide uncertainty intervals that offer reduced actionable value for power grid dispatching. Conversely, PARA-PV avoids this over-conservative behavior, consistently producing narrower prediction intervals across all extended horizons while maintaining reliable coverage boundaries. By minimizing both the normalized interval widths and the average quantile loss across diverse data profiles, the proposed model establishes a robust and practical standard for multi-horizon solar power probabilistic forecasting.

\subsection{Ablation study}

\begin{figure}[!h]
    \centering
    \includegraphics[width=0.86\linewidth]{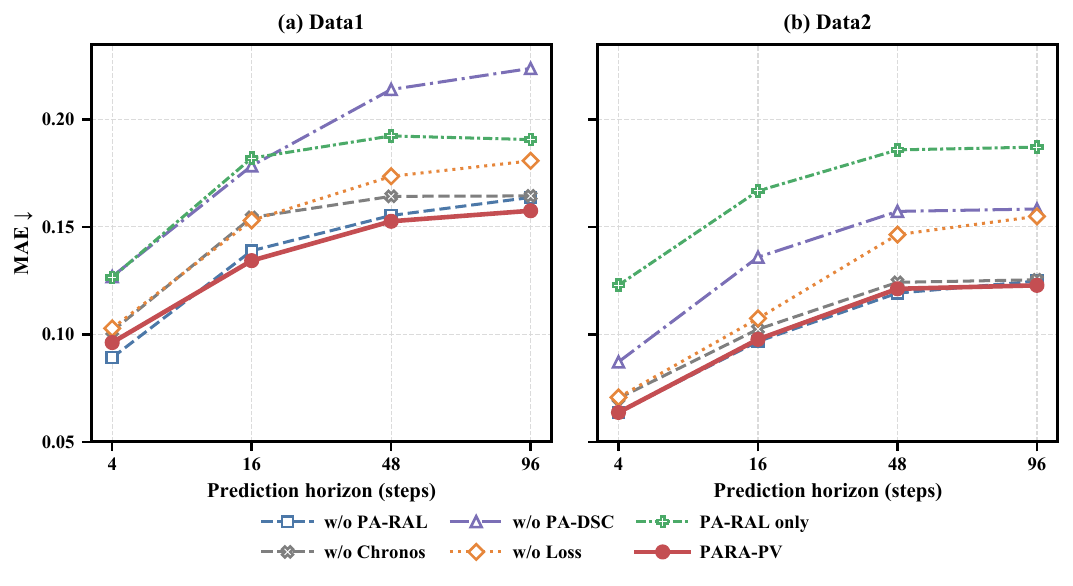}
    \vspace{-0.6\baselineskip}
    \caption{MAE variation across prediction horizons in the ablation study.}
    \label{fig:Data-ablation}
\end{figure}

To thoroughly investigate the individual contribution of each core component within the proposed framework and validate its necessity in driving forecasting performance, we conduct a rigorous ablation study. By systematically disabling specific modules, we can gain a deeper understanding of the explicit benefits each mechanism provides to predictive accuracy and robustness. To this end, we construct five distinct model variants for comparison: (1) w/o PA-RAL, which eliminates the physics-aware mechanism to rely solely on raw retrieval-augmented learner; (2) w/o Chronos, which excludes the frozen time-series foundation model used for prior correction; (3) w/o PA-DSC, which drops the physics-aware distribution shift correction module to evaluate its capability in handling distribution alignment; (4) w/o Loss, which discards the proposed physics-constrained loss function and employs only the traditional MSELoss; and (5) PA-RAL only, which retains PA-RAL and the adaptive segment-balanced loss while excluding Chronos calibration and PA-DSC. The experiment results are presented in \textbf{Fig.~\ref{fig:Data-ablation}}, \textbf{Table~\ref{tab:para_pv_ablation}}, and \textbf{Table~\ref{tab:para_pv_ablation_2}}.

\begin{table*}[!h]
\centering
\captionsetup{width=\textwidth}
\caption{Ablation study of PARA-PV model across multiple prediction horizons in Data 1.}
\label{tab:para_pv_ablation}
\resizebox{1.0\textwidth}{!}{
\begin{tabular}{llcccccc}
\toprule \hline
\multirow{2}{*}{Horizon} 
& \multirow{2}{*}{Metric}
& \multicolumn{6}{c}{Variant} \\
\cmidrule(lr){3-8}
& 
& w/o PA-RAL
& w/o Chronos
& w/o PA-DSC
& w/o Loss
& PA-RAL only
& \textbf{PARA-PV} \\
\midrule

\multirow{4}{*}{4}
& MSE  & \underline{0.0465} & 0.0548 & 0.0480 & 0.0468 & 0.0496 & \textbf{0.0458} \\
& MAE  & \textbf{0.0892} & 0.1015 & 0.1269 & 0.1028 & 0.1265 & \underline{0.0962} \\
& RMSE & \underline{0.2157} & 0.2341 & 0.2192 & 0.2164 & 0.2227 & \textbf{0.2140} \\
& R$^2$ & \underline{0.9386} & 0.9277 & 0.9366 & 0.9382 & 0.9345 & \textbf{0.9396} \\
\midrule

\multirow{4}{*}{16}
& MSE  & \underline{0.0992} & 0.1151 & 0.1012 & 0.1033 & 0.1106 & \textbf{0.0949} \\
& MAE  & \underline{0.1389} & 0.1544 & 0.1784 & 0.1530 & 0.1820 & \textbf{0.1343} \\
& RMSE & \underline{0.3150} & 0.3393 & 0.3181 & 0.3213 & 0.3325 & \textbf{0.3081} \\
& R$^2$ & \underline{0.8685} & 0.8475 & 0.8659 & 0.8632 & 0.8535 & \textbf{0.8742} \\
\midrule

\multirow{4}{*}{48}
& MSE  & 0.1268 & 0.1360 & 0.1409 & \underline{0.1242} & 0.1327 & \textbf{0.1168} \\
& MAE  & \underline{0.1553} & 0.1642 & 0.2139 & 0.1735 & 0.1923 & \textbf{0.1526} \\
& RMSE & 0.3561 & 0.3688 & 0.3754 & \underline{0.3525} & 0.3643 & \textbf{0.3418} \\
& R$^2$ & 0.8317 & 0.8196 & 0.8131 & \underline{0.8352} & 0.8240 & \textbf{0.8450} \\
\midrule

\multirow{4}{*}{96}
& MSE  & 0.1455 & 0.1468 & 0.1581 & 0.1568 & \underline{0.1342} & \textbf{0.1272} \\
& MAE  & \underline{0.1637} & 0.1644 & 0.2237 & 0.1807 & 0.1906 & \textbf{0.1575} \\
& RMSE & 0.3814 & 0.3832 & 0.3976 & 0.3960 & \underline{0.3663} & \textbf{0.3567} \\
& R$^2$ & 0.8062 & 0.8044 & 0.7893 & 0.7911 & \underline{0.8212} & \textbf{0.8305} \\
\midrule

\rowcolor{mediumgray}
& MSE $\downarrow$   & \underline{0.1045} & 0.1132 & 0.1121 & 0.1078 & 0.1068 & \textbf{0.0962} \\
\rowcolor{mediumgray}
& MAE $\downarrow$   & \underline{0.1368} & 0.1461 & 0.1857 & 0.1525 & 0.1729 & \textbf{0.1352} \\
\rowcolor{mediumgray}
& RMSE $\downarrow$  & \underline{0.3170} & 0.3314 & 0.3276 & 0.3216 & 0.3215 & \textbf{0.3052} \\
\rowcolor{mediumgray}
\multirow{-4}{*}{\textbf{AVG}} & R$^2$ $\uparrow$   & \underline{0.8613} & 0.8498 & 0.8512 & 0.8569 & 0.8583 & \textbf{0.8723} \\
\hline

\bottomrule
\end{tabular}
}
\vspace{4pt}
\begin{minipage}{\linewidth}
\raggedright
\footnotesize{
Note: The best results are shown in \textbf{BOLD}, and the second-best results are \underline{UNDERLINED}.
}
\end{minipage}
\end{table*}

\begin{table*}[!h]
\centering
\captionsetup{width=\textwidth}
\caption{Ablation study of PARA-PV model across multiple prediction horizons in Data 2.}
\label{tab:para_pv_ablation_2}
\resizebox{1.0\textwidth}{!}{
\begin{tabular}{llcccccc}
\toprule \hline
\multirow{2}{*}{Horizon} 
& \multirow{2}{*}{Metric}
& \multicolumn{6}{c}{Variant} \\
\cmidrule(lr){3-8}
& 
& w/o PA-RAL
& w/o Chronos
& w/o PA-DSC
& w/o Loss
& PA-RAL only
& \textbf{PARA-PV} \\
\midrule

\multirow{4}{*}{4}
& MSE  & \textbf{0.0274} & 0.0306 & 0.0303 & 0.0281 & 0.0393 & \underline{0.0275} \\
& MAE  & \underline{0.0638} & 0.0703 & 0.0872 & 0.0707 & 0.1229 & \textbf{0.0637} \\
& RMSE & \textbf{0.1656} & 0.1748 & 0.1740 & 0.1676 & 0.1982 & \underline{0.1659} \\
& R$^2$ & \textbf{0.9713} & 0.9680 & 0.9684 & 0.9706 & 0.9589 & \underline{0.9712} \\
\midrule

\multirow{4}{*}{16}
& MSE  & \textbf{0.0597} & 0.0658 & 0.0677 & 0.0616 & 0.0861 & \underline{0.0601} \\
& MAE  & \textbf{0.0967} & 0.1024 & 0.1359 & 0.1074 & 0.1667 & \underline{0.0977} \\
& RMSE & \textbf{0.2443} & 0.2565 & 0.2602 & 0.2483 & 0.2934 & \underline{0.2451} \\
& R$^2$ & \textbf{0.9377} & 0.9312 & 0.9292 & 0.9356 & 0.9101 & \underline{0.9372} \\
\midrule

\multirow{4}{*}{48}
& MSE  & 0.0948 & 0.1008 & 0.0989 & \underline{0.0895} & 0.1155 & \textbf{0.0878} \\
& MAE  & \textbf{0.1192} & 0.1242 & 0.1572 & 0.1464 & 0.1858 & \underline{0.1212} \\
& RMSE & 0.3078 & 0.3175 & 0.3145 & \underline{0.2991} & 0.3398 & \textbf{0.2963} \\
& R$^2$ & 0.9012 & 0.8949 & 0.8969 & \underline{0.9067} & 0.8796 & \textbf{0.9084} \\
\midrule

\multirow{4}{*}{96}
& MSE  & 0.1041 & 0.1043 & 0.1070 & \underline{0.1038} & 0.1224 & \textbf{0.1000} \\
& MAE  & \underline{0.1248} & 0.1254 & 0.1583 & 0.1549 & 0.1871 & \textbf{0.1228} \\
& RMSE & 0.3227 & 0.3229 & 0.3272 & \underline{0.3222} & 0.3499 & \textbf{0.3163} \\
& R$^2$ & 0.8915 & 0.8914 & 0.8885 & \underline{0.8919} & 0.8725 & \textbf{0.8958} \\
\midrule

\rowcolor{mediumgray}
& MSE $\downarrow$   & 0.0715 & 0.0754 & 0.0760 & \underline{0.0708} & 0.0908 & \textbf{0.0689} \\
\rowcolor{mediumgray}
& MAE $\downarrow$   & \textbf{0.1011} & 0.1056 & 0.1346 & 0.1199 & 0.1656 & \underline{0.1014} \\
\rowcolor{mediumgray}
& RMSE $\downarrow$  & 0.2601 & 0.2679 & 0.2690 & \underline{0.2593} & 0.2953 & \textbf{0.2559} \\
\rowcolor{mediumgray}
\multirow{-4}{*}{\textbf{AVG}} & R$^2$ $\uparrow$   & 0.9254 & 0.9214 & 0.9208 & \underline{0.9262} & 0.9053 & \textbf{0.9282} \\
\hline

\bottomrule
\end{tabular}
}
\vspace{4pt}
\begin{minipage}{\linewidth}
\raggedright
\footnotesize{
Note: The best results are shown in \textbf{BOLD}, and the second-best results are \underline{UNDERLINED}.
}
\end{minipage}
\end{table*}

Data 1 demonstrates the distinct impact of each architectural component across different forecasting horizons. The full PARA-PV model achieves the best overall performance, yielding the lowest average MSE of 0.0962 and the highest average $R^2$ of 0.8723. Disabling the time-series foundation model prior correction (w/o Chronos) results in a notable performance drop at the short-term 4-step horizon, where the MSE increases to 0.0548, indicating that foundation model priors are instrumental in capturing immediate temporal patterns. Conversely, for long-term forecasting at 96 steps, omitting the physics-aware distribution shift correction module (w/o PA-DSC) triggers the most severe degradation, causing the MAE to increase to 0.2237. This highlights the indispensable role of the distribution correction module in combating cumulative distribution discrepancies over extended windows. The w/o PA-RAL variant emerges as the second-best model on average, suggesting that while the physics-aware retrieval-augmented learning module is beneficial, its potential is fully realized only when tightly coupled with distribution alignment and structural loss constraints.

Data 2 exhibits a horizon-dependent effect of the PA-RAL design. At the 4- and 16-step horizons, Raw RAL (w/o PA-RAL) marginally outperforms the full model on several metrics. For example, its $R^2$ values are 0.9713 and 0.9377, compared with 0.9712 and 0.9372 for PARA-PV. However, these differences are small and are not consistent across all metrics, since the full model achieves a slightly lower MAE at the 4-step horizon. Therefore, these single-run results do not establish that PA-RAL degrades short-horizon forecasting. At the 48- and 96-step horizons, the full model achieves clearer improvements, suggesting that physics-aware retrieval and analog guidance become more useful as the forecasting horizon increases. Among the degraded variants, the model lacking the physics-constrained loss function (w/o Loss) serves as the closest runner-up, obtaining an average MSE of 0.0708 and an $R^2$ of 0.9262. Meanwhile, the isolated PA-RAL only configuration demonstrates the worst overall metrics, which confirms that individual modules require the cooperative framework of prior corrections and regularized losses to maintain stability.

The empirical results from both datasets indicate a clear complementary relationship among the core modules of the proposed framework. Chronos foundation model priors are critical for short-term precision, while the physics-aware distribution shift correction and specialized loss functions act as regularizers to prevent error propagation over long forecasting horizons. The performance drop of the PA-RAL only variant shows that a single retrieval mechanism is insufficient for complex photovoltaic power regression. Overall, the full PARA-PV model strikes a balance between short-term flexibility and long-term stability across various capacity profiles, confirming the effectiveness of an integrated framework.

\subsection{Computational complexity study}

\begin{figure}[!h]
    \centering
    \includegraphics[width=0.7\linewidth]{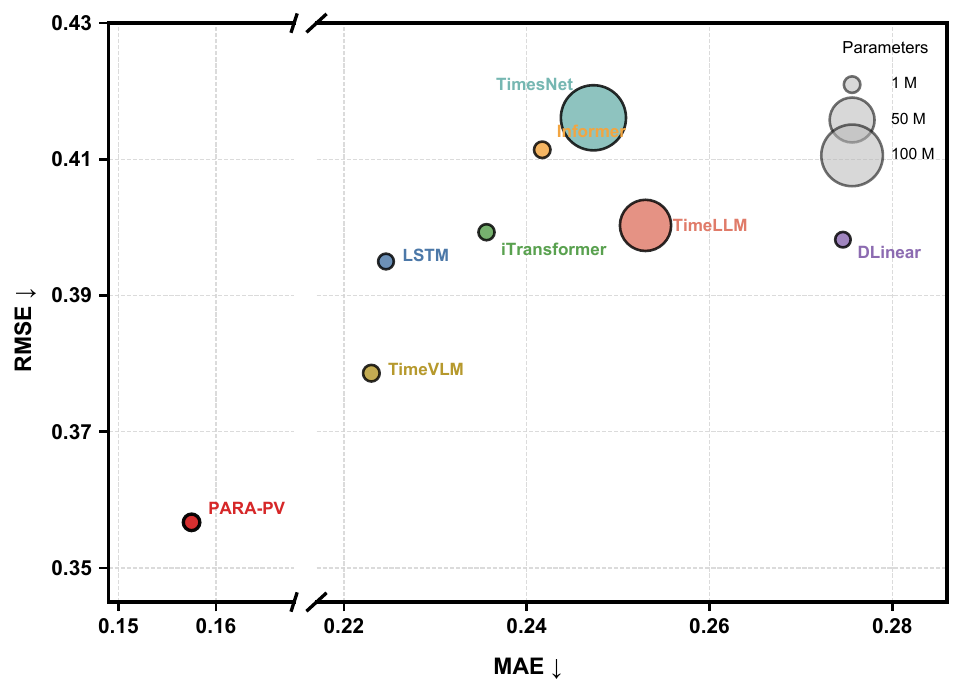}
    \vspace{-0.4\baselineskip}
    \caption{Forecasting performance versus model complexity on Data1 at the 96-Step horizon.}
    \label{fig:Data1-versus}
\end{figure}

To evaluate the computational efficiency of the proposed model, we conduct a complexity analysis using Data 1 under the forecasting task with a prediction horizon of 96. Specifically, we statistically compare the baseline models and the proposed PARA-PV model in terms of training time, inference time, trainable parameters, and peak GPU memory consumption. This setting enables a consistent assessment of the computational overhead introduced by the proposed architecture while reflecting its practical deployment cost in long-horizon PV power forecasting scenarios, as summarized in \textbf{Table~\ref{tab:efficiency_model_comparison}}. \textbf{Fig~\ref{fig:Data1-versus}} further shows the trade-off between forecasting accuracy and model complexity. 

\begin{table*}[!h]
\centering
\footnotesize
\captionsetup{width=\textwidth}
\caption{Computational efficiency comparison of baseline models and PARA-PV.}
\label{tab:efficiency_model_comparison}
\setlength{\textwidth}{7pt}
\renewcommand{\arraystretch}{1.2}
\begin{tabular}{lcccc}
\toprule \hline
\multirow{2}{*}{Model}
& \multicolumn{4}{c}{Efficiency Metrics} \\
\cmidrule(lr){2-5}
& $T_{\mathrm{train}}$ $\downarrow$
& $T_{\mathrm{inf}}$ $\downarrow$
& Params $\downarrow$
& Mem. $\downarrow$ \\
\midrule
LSTM & 6.4791 & 0.00000044 & 231.39K & 89.09MB \\
Informer & 22.6946 & 0.00000211 & 915.33K & 77.84MB \\
DLinear & 4.2642 & 0.00000013 & 37.06K & 17.92MB \\
iTransformer & 12.8515 & 0.00000067 & 565.47K & 24.46MB \\
TimesNet & 971.8936 & 0.00012840 & 112529.82K & 1329.56MB \\
TimeLLM & 268.3760 & 0.00007062 & 66991.02K & 3450.58MB \\
TimeVLM & 98.2159 & 0.00002826 & 1081.37K & 754.09MB \\
\rowcolor{mediumgray}\textbf{PARA-PV} & 201.5256 & 0.00006902 & 993.22K & 681.52MB \\
\hline
\bottomrule
\end{tabular}
\vspace{4pt}
\begin{minipage}{\linewidth}
\raggedright
\footnotesize{Note: $T_{\mathrm{train}}$ denotes the training time per epoch, measured in s/epoch; $T_{\mathrm{inf}}$ denotes the average inference time per predicted step and is calculated as$T_{\mathrm{inf}}
=\frac{\sum_{b=1}^{B_{\mathrm{test}}}\Delta t_b}
{N_{\mathrm{test}}H},$
where $\Delta t_b$ is the synchronized wall-clock inference time of the $b$-th test batch, $N_{\mathrm{test}}$ is the number of test samples, and $H$ is the prediction horizon. Params denotes the number of trainable parameters, and Mem. denotes peak GPU memory usage. Lower values indicate better computational efficiency.}
\end{minipage}
\end{table*}

The proposed PARA-PV strikes a competitive trade-off between large-scale forecasting architectures and lightweight baselines in terms of computational efficiency. Compared to large models like TimesNet and TimeLLM, PARA-PV substantially reduces trainable parameters to 993.22K, compared to 112529.82K for TimesNet and 66991.02K for TimeLLM. This parameter efficiency lowers the peak GPU memory to 681.52MB, approximately half that of TimesNet (1329.56MB) and less than one-fifth of TimeLLM (3450.58MB). PARA-PV also requires 201.5256 s/epoch for training and 0.00006902 s/step for inference, outperforming both models in execution speed. Although TimeVLM trains faster (98.2159 s/epoch) and infers quicker (0.00002826 s/step), PARA-PV maintains a smaller parameter size and lower memory footprint than TimeVLM (1081.37K parameters and 754.09MB memory). In contrast, compared to lightweight baselines such as DLinear, iTransformer, Informer, and LSTM, PARA-PV incurs higher computational overhead due to its more complex architecture designed for long-horizon forecasting. For instance, DLinear requires only 37.06K parameters, 4.2642 s/epoch training time, and 17.92MB peak memory. However, by keeping its total parameters under 1M, PARA-PV avoids excessive resource growth, ensuring practical feasibility for deployment.

\subsection{Sensitivity study}

This section analyzes the sensitivity of the proposed PARA-PV framework to key hyper-parameters. We selected three representative hyper-parameters, including learning rate (LR), dropout rate (DR), and PA-RAL patch configuration (PC). Here, PC denotes the patch length and stride used in the physics-aware retrieval branch. The candidate settings are LR = \{0.00005, 0.0001$^{*}$, 0.00015, 0.0002\}, DR = \{0.05, 0.1$^{*}$, 0.2, 0.3\}, and PC = \{(8,4), (16$^{*}$,8$^{*}$), (24,12), (32,16)\}. The asterisk indicates the adopted best setting. LR$_{\phi}$(DR, PC) denotes changing LR while keeping DR and PC unchanged; DR$_{\phi}$(LR, PC) and PC$_{\phi}$(LR, DR) are defined similarly. We report MSE, MAE, RMSE, and $R^2$ under different settings, and use the standard deviation (STD) to measure the sensitivity of PARA-PV to each hyper-parameter. The experimental results are summarized in \textbf{Table~\ref{tab:sensitivity_data1}}. For a given evaluation metric under $K$ candidate settings of a hyper-parameter, the STD is calculated as follows:
\begin{equation}
    \mathrm{STD}
    =
    \sqrt{
    \frac{1}{K}
    \sum_{k=1}^{K}
    \left(m_k - \bar{m}\right)^2
    },
    \quad
    \bar{m}
    =
    \frac{1}{K}
    \sum_{k=1}^{K} m_k,
\end{equation}
where $m_k$ denotes the metric value obtained under the $k$-th hyper-parameter setting, and $\bar{m}$ is the mean value across all candidate settings.

\begin{table}[!h]
\centering
\caption{Sensitivity analysis results of PARA-PV.}
\label{tab:sensitivity_data1}
\resizebox{\textwidth}{!}{
\begin{tabular}{llcccccc}
\toprule
\multirow{2}{*}{Dataset} & \multirow{2}{*}{Metric}
& \multicolumn{3}{c}{16-step}
& \multicolumn{3}{c}{96-step} \\
\cmidrule(lr){3-5} \cmidrule(lr){6-8}
& & LR$_{\phi}$(DR, PC)
& DR$_{\phi}$(LR, PC)
& PC$_{\phi}$(LR, DR)
& LR$_{\phi}$(DR, PC)
& DR$_{\phi}$(LR, PC)
& PC$_{\phi}$(LR, DR) \\
\midrule
\multirow{4}{*}{Data 1}
& MSE  & 0.001728 & 0.003932 & 0.001834 & 0.001355 & 0.001265 & 0.000957 \\
& MAE  & 0.001623 & 0.005742 & 0.002337 & 0.001965 & 0.002078 & 0.001674 \\
& RMSE & 0.002770 & 0.006186 & 0.002927 & 0.001871 & 0.001748 & 0.001335 \\
& R$^2$ & 0.002289 & 0.005210 & 0.002429 & 0.001805 & 0.001685 & 0.001274 \\
\midrule
\multirow{4}{*}{Data 2}
& MSE  & 0.001441 & 0.002280 & 0.000455 & 0.002725 & 0.002576 & 0.000913 \\
& MAE  & 0.000957 & 0.005075 & 0.001108 & 0.002102 & 0.004867 & 0.000241 \\
& RMSE & 0.002859 & 0.004503 & 0.000911 & 0.004248 & 0.004000 & 0.001426 \\
& R$^2$ & 0.001505 & 0.002382 & 0.000476 & 0.002838 & 0.002683 & 0.000951 \\

\bottomrule
\end{tabular}
}
\end{table}

The sensitivity analysis demonstrates the inherent robustness of the proposed PARA-PV framework, as reflected by the consistently low STD values across all parameter variations. Across both datasets and both forecasting horizons, the performance fluctuations induced by altering the LR, DR, and PC remain minimal, with all reported STD values staying well below 0.0062. Even the maximum variance observed, which occurs for the DR under the 16-step horizon on Data 1, yields an RMSE STD of only 0.006186, indicating that the model performance remains tightly bounded. For the long-term forecasting task (96-step), this insensitivity becomes even more pronounced, where the STDs for MSE across all three hyperparameters span a narrow and low range between 0.000913 and 0.002725. The structural parameter PC exhibits the highest stability, achieving an exceptionally low MAE STD of 0.000241 on Data 2 under the 96-step setting. These uniformly small metrics across different evaluation dimensions, datasets, and horizons demonstrate that PARA-PV is highly insensitive to hyperparameter choices, mitigating the need for exhaustive fine-tuning and ensuring stable deployment in practical scenarios.

\section{Conclusion}

This paper introduces PARA-PV, a physics-aware retrieval-augmented framework for multi-step PV power forecasting. The framework addresses the challenges of regime-biased and physically inconsistent predictions caused by weather variability, diurnal cycles, and complex physical constraints. The framework integrates four primary components. First, a physics-aware retrieval-augmented learner (PA-RAL) uses patch-level temporal embeddings to capture historical analog contexts from a memory bank. Second, a Chronos-guided temporal prior calibration module incorporates the broad temporal knowledge of a frozen foundation model into local PV constraints via a lightweight residual adapter. Third, a physics-aware distribution shift correction (PA-DSC) module employs a dilated temporal convolutional network to reduce post-prediction bias induced by shifting environmental and temporal conditions. Finally, a physics-constrained loss function balances optimization weights across distinct operational regimes. Deterministic and probabilistic forecasting experiments on real-world datasets show that PARA-PV consistently outperforms representative baseline models. The results demonstrate that the framework achieves high prediction accuracy, reliable interval calibration under uncertainty, and strong generalization across different horizons, while maintaining a low computational burden.

In practical deployment, PARA-PV can support power grid dispatch, energy storage management, and electricity market operations. By providing reliable forecasts of solar generation trends and rapid ramping behaviors, the model helps operators optimize resource allocation, improve grid acceptance of volatile renewable energy, and enhance overall system reliability. However, several limitations remain. The current retrieval mechanism relies on a static memory bank, restricting its adaptability to long-term equipment degradation and environmental changes. Additionally, the framework has only been evaluated on solar power data. Future work will address these deployment challenges in two directions. We plan to develop an online continuous learning paradigm with a dynamic memory update mechanism to incorporate real-time system feedback without full offline retraining. We also intend to extend the physics-aware retrieval and segment-balanced optimization architectures to wind energy and load forecasting, providing a generalized forecasting solution for multi-energy microgrid management.

\section*{Data availability}

We have utilized public datasets.

\section*{Acknowledgment}

This work was supported by the Natural Science Foundation of Hebei Province (G2025502003), by the Research on Electricity Price Forecasting and Intelligent Bidding Strategy of Compressed Air Energy Storage in Qinghai Spot Market (KFKT-26LAB-01), and by the Fundamental Research Funds for the Central Universities (2025JC002).

\appendix
\section{Parameter settings}
\label{appendix.A}

To improve the reproducibility of the experiments, we summarize the key hyperparameters used in this study in two tables. \textbf{Table~\ref{tab:hyperparameters}} reports the main hyperparameters of the proposed PARA-PV framework, including the common experimental settings, backbone representation parameters, PA-RAL retrieval configuration, Chronos prior calibration settings, PA-DSC correction parameters, and loss function. During each forward pass, representations and associated metadata from all channels in the current mini-batch are inserted into the PA-RAL memory bank. The memory bank is continuously updated online under a fixed-capacity constraint. Once its capacity is reached, incoming items overwrite the oldest stored entries through a circular replacement strategy, maintaining an up-to-date collection of recent patterns.

\textbf{Table~\ref{tab:baseline_hyperparameters}} lists the corresponding hyperparameters of the baseline models used for comparison, covering both shared training settings and model-specific configurations. These settings are kept consistent across forecasting horizons whenever possible, so that the experimental comparison mainly reflects the modeling capability of each method rather than differences in training protocols.

\begin{table}[!h]
\centering
\caption{Key parameters of PARA-PV.}
\label{tab:hyperparameters}
\begin{tabular}{lll}
\toprule
Module & Parameter & Value \\
\midrule
                     & Input length $L$ & 192 \\
                     & Prediction horizon $H$ & $\{4,16,48,96\}$ \\
                     & Number of input variables & 7 \\
Experimental setting & Batch size & 32 \\
                     & Training epochs & 50 \\
                     & Learning rate & 0.001 \\
                     & Dropout & 0.1 \\
\midrule
                        & Hidden dimension $d_{\mathrm{model}}$ & 128 \\
                        & Number of attention heads & 8 \\
Backbone representation & Encoder layers & 2 \\
                        & Feed-forward dimension & 768 \\
                        & Normalization constant & 0.4 \\
\midrule
       & Patch length & 16 \\
       & Patch stride & 8 \\
PA-RAL & Patch memory size & $\{100,512,1024,2048,4096^*,8192\}$ \\
       & Retrieved neighbors $K$ & 5 \\
       & Intra-day buckets & 24 \\
\midrule
 \multirow{4}{*}{Chronos prior calibration}  & Chronos quantile & 0.5 \\
                    & Adapter hidden dimension & 96 \\
                    & Adapter dropout & 0.1 \\
                    & Version & Chronos-2 120M \\
\midrule
\multirow{2}{*}{PA-DSC}   & Hidden dimension & 128 \\
                          & Low-power margin & 0.1 \\
\midrule
Loss function & Quantile levels & $\{0.05,0.10,0.50,0.90,0.95\}$ \\
\bottomrule
\end{tabular}
\vspace{4pt}
\begin{minipage}{\linewidth}
\raggedright
\footnotesize{Note: * indicates the optimal parameter configuration.}
\end{minipage}
\end{table}

\begin{table}[!h]
\centering
\caption{Key parameters of baseline models.}
\label{tab:baseline_hyperparameters}
\begin{tabular}{lll}
\toprule
Model & Parameter & Value \\
\midrule
              & Input length $L$ & 192 \\
              & Prediction horizon $H$ & $\{4,16,48,96\}$ \\
\multirow{2}{*}{All baselines}  & Batch size & 32 \\
              & Training epochs & 50 \\
              & Learning rate & 0.001 \\
              & Dropout & 0.1 \\
\midrule
\multirow{2}{*}{LSTM} & Hidden dimension & 128 \\
                      & Number of recurrent layers & 2 \\
\midrule
         & Model dimension $d_{\mathrm{model}}$ & 128 \\
         & Number of attention heads & 8 \\
\multirow{2}{*}{Informer} & Encoder layers & 2 \\
         & Decoder layers & 1 \\
         & Feed-forward dimension & 768 \\
         & Attention factor & 1 \\
\midrule
\multirow{2}{*}{DLinear} & Decomposition kernel size & 25 \\
         & Individual linear layers & False \\
\midrule
             & Model dimension $d_{\mathrm{model}}$ & 128 \\
\multirow{2}{*}{iTransformer} & Number of attention heads & 8 \\
             & Encoder layers & 2 \\
             & Feed-forward dimension & 768 \\
\midrule
         & Model dimension $d_{\mathrm{model}}$ & 128 \\
         & Number of TimesBlocks & 2 \\
TimesNet & Top-$k$ periods & 5 \\
         & Number of convolution kernels & 6 \\
         & Feed-forward dimension & 768 \\
\midrule
        & LLM backbone & GPT-2 \\
        & LLM hidden dimension & 768 \\
TimeLLM & Number of LLM layers & 1 \\
        & Patch length & 16 \\
        & Patch stride & 8 \\
\midrule
        & VLM backbone & CLIP \\
        & Image size & 224 \\
\multirow{2}{*}{TimeVLM} & Patch length & 16 \\
        & Patch stride & 8 \\
        & Patch memory size & 100 \\
        & Learnable image mapping & True \\
\bottomrule
\end{tabular}
\end{table}

\clearpage
\bibliographystyle{elsarticle-num} 
\bibliography{Reference}

\end{document}